\documentclass[journal]{IEEEtran}%
\usepackage{graphicx}%
\usepackage{amsmath}%
\usepackage{amssymb}%
\usepackage{subfig}%
\usepackage[table]{xcolor}%
\usepackage{hyperref}%
\hypersetup{%
colorlinks,%
linkcolor={red!50!black},%
citecolor={blue!50!black},%
urlcolor={blue!80!black}%
}%
\usepackage{doi}%

\global\let\myhref\href%
\global\let\mydoi\doi%
\protected\gdef\objFun{\ensuremath{f}}%
\protected\gdef\objFunb#1{\mbox{\ensuremath{\objFun\!\left(#1\right)}}}%
\protected\gdef\objectiveSpace{\ensuremath{\mathcal{Y}}}%
\protected\gdef\obspel{\ensuremath{y}}%
\protected\gdef\obspeli#1{\mbox{\ensuremath{\obspel_{#1}}}}%
\protected\gdef\scale{\ensuremath{s}}%
\protected\gdef\upperBound{\mbox{\ensuremath{U\!B}}}%
\protected\gdef\solutionSpace{\ensuremath{\mathcal{X}}}%
\protected\gdef\solspel{\ensuremath{x}}%
\protected\gdef\solspeli#1{\mbox{\ensuremath{\solspel_{#1}}}}%
\protected\gdef\solspelval#1{\mbox{\ensuremath{\solspel\!\left[#1\right]}}}%
\protected\gdef\solspels#1{\mbox{\ensuremath{\solspel^{(#1)}}}}%
\protected\gdef\solspelps#1{\mbox{\ensuremath{\hat{\solspel}^{(#1)}}}}%
\protected\gdef\ffaH{\ensuremath{H}}%
\protected\gdef\ffaHb#1{\mbox{\ensuremath{\ffaH\!\left[#1\right]}}}%
\definecolor{algobg}{HTML}{EEEEEE}%
\definecolor{opoea}{HTML}{FFA500}%
\protected\gdef\opoeaP{\mbox{$(\!1\!\!+\!\!1\!)$\hspace*{0.2em}EA}}%
\protected\gdef\opoeaC#1{{\color{opoea}{#1}}}%
\protected\gdef\opoea{\opoeaC{\opoeaP}}%
\definecolor{opofea}{HTML}{E82000}%
\protected\gdef\opofeaC#1{{\color{opofea}{#1}}}%
\protected\gdef\opofeaP{\mbox{$(\!1\!\!+\!\!1\!)$\hspace*{0.2em}FEA}}%
\protected\gdef\opofea{\opofeaC{\opofeaP}}%
\definecolor{saga}{HTML}{4DE16B}%
\protected\gdef\sagaC#1{{\color{saga}{#1}}}%
\protected\gdef\ollga{(1+($\lambda$,$\lambda$))~GA}%
\protected\gdef\sagaP{\mbox{SAGA}}%
\protected\gdef\saga{\sagaC{\sagaP}}%
\protected\gdef\sagaL{Self-Adjusting~\mbox{(1+($\lambda$,$\lambda$))~GA}}%
\definecolor{safga}{HTML}{018F61}%
\protected\gdef\safgaC#1{{\color{safga}{#1}}}%
\protected\gdef\safgaP{\mbox{SAFGA}}%
\protected\gdef\safga{\safgaC{\safgaP}}%
\definecolor{gga}{HTML}{8B66FF}%
\protected\gdef\ggaC#1{{\color{gga}{#1}}}%
\protected\gdef\ggaP{\mbox{GGA}}%
\protected\gdef\gga{\ggaC{\ggaP}}%
\protected\gdef\ggaL{\mbox{Greedy~(2+1)~GA}}%
\definecolor{gfga}{HTML}{952659}%
\protected\gdef\gfgaC#1{{\color{gfga}{#1}}}%
\protected\gdef\gfgaP{\mbox{GFGA}}%
\protected\gdef\gfga{\gfgaC{\gfgaP}}%
\definecolor{safgap}{HTML}{0000FF}%
\protected\gdef\safgapC#1{{\color{safgap}{#1}}}%
\protected\gdef\safgapP{\mbox{SAHGA}}%
\protected\gdef\safgap{\safgapC{\safgapP}}%
\definecolor{eafea}{HTML}{00CCCC}%
\protected\gdef\eafeaC#1{{\color{eafea}{#1}}}%
\protected\gdef\eafeaP{\mbox{EAFEA}}%
\protected\gdef\eafea{\eafeaC{\eafeaP}}%
\protected\gdef\countones#1{\mbox{\ensuremath{\left|#1\right|_1}}}%
\gdef\onemax{\mbox{OneMax}}%
\gdef\trap{\mbox{Trap}}%
\gdef\leadingones{\mbox{LeadingOnes}}%
\gdef\twomax{\mbox{TwoMax}}%
\gdef\jump{\mbox{Jump}}%
\protected\gdef\jumpWidth{\mbox{\ensuremath{w}}}%
\gdef\plateau{\mbox{Plateau}}%
\global\let\plateauWidth\jumpWidth%
\gdef\nqueensN{\mbox{N}}%
\gdef\nqueens{\mbox{\nqueensN\hspace{-0.03em}-\hspace{-0.03em}Queens}}%
\gdef\ising{\mbox{Ising}}%
\gdef\isingod{\mbox{\ising\ 1D}}%
\gdef\isingtd{\mbox{\ising\ 2D}}%
\gdef\linearHarmonic{\mbox{linHarm}}%
\gdef\wmodel{\mbox{W-Model}}%
\gdef\maxsat{\mbox{MaxSat}}%
\protected\gdef\runtime{\mbox{\textnormal{RT}}}%
\protected\gdef\ert{\mbox{\textnormal{E\runtime}}}%
\protected\gdef\bigO{\ensuremath{\mathbf{O}}}%
\protected\gdef\bigOof#1{\mbox{\ensuremath{\bigO\!\left(#1\right)}}}%
\protected\gdef\bigOmegaOf#1{\mbox{\ensuremath{{\mathbf \Omega}\!\left(#1\right)}}}%
\protected\gdef\bigThetaOf#1{\mbox{\ensuremath{{\mathbf \Theta}\!\left(#1\right)}}}%
\protected\gdef\npPrefix{\ensuremath{\mathcal{NP}}}%
\protected\gdef\npHard{\mbox{\npPrefix-hard}}%
\protected\gdef\booleans{\mbox{\ensuremath{\{\hspace{-0.1em}0,1\hspace{-0.1em}\}}}}%
\protected\gdef\bestSoFarX{\solspeli{\textnormal{B}}}%
\protected\gdef\bestSoFarF{\obspeli{\textnormal{B}}}%
\protected\gdef\maxSatClauses{\ensuremath{c}}%
\global\let\maxSatVariables\scale%
\protected\gdef\maxSatFormula{\mbox{\ensuremath{B}}}%
\protected\gdef\intFromTo#1#2{\ensuremath{[#1..#2]}}%
\def\mathrlap{\mathpalette\mathrlapinternal}%
\def\mathrlapinternal#1#2{\rlap{$\mathsurround=0pt#1{#2}$}}%
\begin{document}
\author{Thomas~Weise, Zhize~Wu, Xinlu~Li, Yan~Chen, and J{\"o}rg L{\"a}ssig%
\thanks{%
T.~Weise, Z.~Wu, X.~Li, and Y.~Chen are with the %
Institute of Applied Optimization, %
School of Artificial Intelligence and Big Data, %
Hefei University, %
Hefei, Anhui, China~230601. %
J.~L{\"a}ssig is with the %
Enterprise Application Development Group of the %
Faculty of Electrical Engineering and Computer Science of the %
Hochschule Zittau/G{\"o}rlitz,
G{\"o}rlitz, Germany. %
Corresponding author: Zhize~Wu, \myhref{mailto:wuzhize@mail.ustc.edu.cn}{wuzhize@mail.ustc.edu.cn}%
}}%
\title{Frequency Fitness Assignment: Optimization without Bias for Good Solutions can be Efficient}%
\markboth{%
IEEE Transactions on Evolutionary Computation,~Vol.~XX, No.~XX, XX~XXXX
}{%
Weise et al.,%
Frequency Fitness Assignment 3%
}%
\maketitle%
\begin{abstract}%
A fitness assignment process transforms the features (such as the objective value) of a candidate solution to a scalar fitness, which then is the basis for selection.
Under Frequency Fitness Assignment~(FFA), the fitness corresponding to an objective value is its encounter frequency in selection steps and is subject to minimization.
FFA creates algorithms that are not biased towards better solutions and are invariant under all injective transformations of the objective function value.
We investigate the impact of FFA on the performance of two theory-inspired, state-of-the-art EAs, the \ggaL\ and the \sagaL.
FFA improves their performance significantly on some problems that are hard for them.
In our experiments, one FFA-based algorithm exhibited mean runtimes that appear to be polynomial on the theory-based benchmark problems in our study, including traps, jumps, and plateaus.
We propose two hybrid approaches that use both direct and FFA-based optimization and find that they perform well.
All FFA-based algorithms also perform better on satisfiability problems than any of the pure algorithm variants.%
\end{abstract}%
\let\oldtheffotnote\thefootnote%
\let\thefootnote\relax\footnote{%
This is revision~1 of the paper that has been submitted for review to the IEEE Transactions on Evolutionary Computation.
\copyright\ 2022 IEEE. Personal use of this material is permitted. Permission from IEEE must be obtained for all other uses, in any current or future media, including reprinting/republishing this material for advertising or promotional purposes, creating new collective works, for resale or redistribution to servers or lists, or reuse of any copyrighted component of this work in other works.}%
\let\thefootnote\oldtheffotnote%
\setcounter{footnote}{0}%
\section{Introduction}%
\IEEEPARstart{F}{itness} assignment is a component of many Evolutionary Algorithms~(EAs).
It transforms the features of candidate solutions, such as their objective value(s), to scalar values which are then the basis for selection. 
Usually, the goal is to maintain diversity in the population in order to avoid getting stuck at local optima.
Under Frequency Fitness Assignment~(FFA), the fitness of a candidate solution is the absolute encounter frequency of its objective value so far during the optimization process~\cite{WWTWDY2014FFA}.
Being subject to minimization, FFA drives the search away from already-discovered objective values and towards solutions with new qualities.

FFA breaks with one of the most fundamental concept of heuristic optimization:
FFA-based algorithms are not biased towards better solutions~\cite{WLCW2021SJSSPWUABFGS}, i.e., they do not prefer better solutions over worse ones.
They also are invariant under all injective transformations of the objective function value, which is the strongest invariance property of any non-trivial single-objective optimization algorithm~\cite{WWLC2020FFAMOAIUBTOTOFV}.\footnote{%
We stated in~\cite{WWLC2020FFAMOAIUBTOTOFV} that FFA makes algorithms invariant under \emph{bijections} of the objective function value. Yet, \emph{surjectivity} is not needed, \emph{injectivity} is already sufficient. We thank an anonymous reviewer for finding this.%
} %
Only random sampling, random walks, and exhaustive enumeration have similar properties and neither of them is considered to be an efficient optimization method.

One would expect that this comes at a significant performance penalty.
Yet, FFA performed well in Genetic Programming tasks with their often rugged, deceptive, and highly epistatic landscapes~\cite{WWTY2014EEIAWGP,WWTWDY2014FFA} and on a benchmark problem simulating such landscapes~\cite{WCLW2019SADSOBIFATMPFBBDOA}.
While the~\opoeaP\ has exponential expected runtime on problems such as \jump, \twomax, and \trap, the \opofeaP, the same algorithm but using FFA, exhibits mean runtimes that appear to be polynomial in experiments and also solves \maxsat\ problems much faster than the \opoeaP~\cite{WWLC2020FFAMOAIUBTOTOFV}.

These interesting properties and results lead to the question whether FFA could also benefit state-of-the-art black-box metaheuristics.
In this article, we investigate the behavior of FFA when plugged into two such algorithms, the \ggaL\ (\ggaP)~\cite{S2012CSUBBA,S2017HCSUBBAIGA} and the \sagaL\ (\sagaP)~\cite{DDE2015FBBCTDNGA,DD2018OSASAPCFT1LLGA}.
\ggaP\ is the first GA with a provable guarantee to be significantly faster than any mutation-only GA and \sagaP\ is the first GA with linear expected runtime, both on the well-known \onemax\ problem (see Section~\ref{sec:onemax}).

We conduct a series of experiments applying both EAs (in their efficient modified form from~\cite{CPD2017TAMPARAOEA}) with and without FFA to a wide range on well-known benchmark problems from discrete optimization theory~\cite{DYHWSB2020BDOHWI}.
We observe that FFA leads to a performance decrease on problems that the algorithms already can solve well.
On some of the problems that they cannot solve efficiently, FFA can provide a significant performance improvement, e.g., reducing exponential to empirical polynomial mean runtime.
To investigate whether FFA can be combined with traditional optimization to reap the best performance in all scenarios, we develop two hybrid algorithms, the \eafeaP, which executes the steps of the \opoeaP\ and \opofeaP\ in an alternating fashion, and the \safgapP, a \sagaP\ variant that can switch between direct and FFA-based optimization.

The eight core contributions of this article are as follows:

\textbf{1)}~We develop FFA-based variants of two state-of-the-art EAs for discrete optimization.

\textbf{2)}~We develop two hybrid algorithms that use both the objective function directly and FFA.

\textbf{3)}~We investigate the performance of these algorithms in a very large-scale study on eleven theory-inspired benchmarks, some of which invoke exponential runtime of the base algorithms, and the satisfiability problem.
We conduct more than 56~million runs, consuming more than 150~processor years, to ensure that our results are statistically sound and rigorous.

\textbf{4)}~When plugged into a \ggaP\ or \opoeaP, FFA can provide runtime that appears to be polynomial in our experiments on \trap, \twomax, and \jump, where these algorithms otherwise require exponential expected runtime.
It does not improve their exponential expected runtime on \plateau\ problems.

\textbf{5)}~The \safgaP, i.e., the \sagaP\ with FFA, as well as the hybrid \safgapP\ have empirical polynomial runtime on \emph{all} of the investigated theory-inspired benchmark problems, including {\plateau}s.

\textbf{6)}~On the \maxsat\ problem, \emph{every} FFA-based algorithm outperforms \emph{all} of the pure methods by a huge margin.

\textbf{7)}~We confirm that our hybrid algorithms exhibit good performance.

\textbf{8)}~With this, we have substantiated the high attractiveness and also practical utility of FFA.
FFA is the only approach that allows for efficient optimization without bias for better solutions.
This makes it also highly interesting from a theoretical point of view.

All experimental results, all source codes, as well as all the code that creates the diagrams and tables from the results are available in an immutable archive at~\mydoi{10.5281/zenodo.5567725}.

The remainder of this paper is structured as follows:
Section~\ref{sec:ffa} discusses FFA and in Section~\ref{sec:algorithms}, we introduce the algorithms studied.
The experimental setup and results are presented in Section~\ref{sec:experiments} and Section~\ref{sec:conclusions} summarizes our findings and plans for future work.%
\section{Frequency Fitness Assignment}%
\label{sec:ffa}%
The objective function~$\objFun:\solutionSpace\rightarrow\objectiveSpace$ maps each candidate solution~$\solspel\in\solutionSpace$ to an objective value~$\obspel=\objFunb{\solspel}$.
Optimization algorithms normally use the objective values to decide which solution should be further explored.
To improve the diversity in populations of EAs, fitness assignment processes like sharing and niching~\cite{GR1987GAS} combine the objective values and density information into a fitness value.
The EA then uses the fitness instead of the objective value as basis for selection.

FFA is a fitness assignment process suitable for problems where the number~$|\objectiveSpace|$ of possible objective values is not too high.
This is true for many important hard problems, like \maxsat, set covering, job shop scheduling, and vertex covering, for many well-studied benchmark problems in discrete optimization, as well as for several machine learning tasks, such as classification~\cite{WWTWDY2014FFA} or object detection~\cite{WWZXLW2021HODFVHRSI}.

Under FFA, the fitness of a candidate solution~\solspel\ is the encounter frequency~\ffaHb{\obspel} of its objective value~$\obspel=\objFunb{\solspel}$ so far during the search.
The selection step of a metaheuristic chooses which candidate solutions to retain and which to remove from the population.
Before selection, we increment the fitness corresponding to the objective value of each solution in the population by one.
If one objective value occurs multiple times, its fitness is incremented multiple times as well.
The fitness is subject to minimization and, as the sole criterion used for selection, replaces the objective value in all comparisons.\footnote{%
The search will, of course, remember the best-so-far solution based on the objective value in a special variable to determine its final result, but this variable is not used in any selection decision, see Figure~\ref{fig:algorithmsA}.%
}

Whenever a solution~\solspeli{n} with a previously unseen objective value~$\obspeli{n}=\objFunb{\solspeli{n}}$ is encountered, it will receive the best possible fitness~$\ffaHb{\obspeli{n}}=1$ since its fitness is initially zero and then incremented to one before selection.
Thus, any new best-so-far solution will also receive the best fitness under FFA when first encountered.
FFA turns static problems into dynamic ones:
Every time a solution with the same objective value~\obspeli{n} is encountered or retained in the population, $\ffaHb{\obspeli{n}}$ is increased.
This makes them less and less attractive and the search will eventually depart from them.

Counting the occurrences of the results of~\objFunb{\solspel} or of~$g(\objFunb{\solspel})$ yields the same algorithm behavior as long as~$g$ maps each unique value of~\objFun\ to a unique value.
Optimization algorithms basing their decisions on the frequency fitness alone therefore are invariant under any injective function~$g$ applied to the objective values~($g\circ\objFun$)~\cite{WWLC2020FFAMOAIUBTOTOFV}, including scaling, shifting, permutations, and even \emph{encryption}.
For the same reason, they are not biased towards better solutions.
They will accept a new solution~\solspeli{n} as long as its objective value has an encounter frequency not higher than the one of the current solution~\solspeli{c}, regardless whether it is better or worse.
That such a scheme can solve problems efficiently is surprising, since preferring better solutions over worse ones (with a certain probability) is maybe the most central concept in optimization.

The related works on FFA are discussed in detail in our recent paper~\cite{WWLC2020FFAMOAIUBTOTOFV}.
Invariance was one major design principle of Information-Geometric Optimization~\cite{OAAH2017IGOAAUPVIP}, which is maybe the closest related work in terms of invariance properties.
Under FFA, an algorithm may oscillate between pursuing increasing and decreasing objective values.
The idea of inverse tournament selection~\cite{CLOW2021OSSEAASPWIRBAORPIB}, where the loser is selected, is interesting in this context.
FFA might also be considered as a Quality-Diversity approach~\cite{GLY2019QDTS,CD2018QADOAUMF,PSS2016QDANFFEC} like Novelty Search~\cite{LS2011AOETTSFNA} or Surprise Search~\cite{GLY2016SSBOAN}.
However, different from all of these methods, FFA itself is not an algorithm but a module that can be plugged into existing algorithms.%
\section{Investigated Algorithms}%
\begin{figure*}%
\centering\noindent%
\subfloat[\opoea~\cite{CPD2017TAMPARAOEA,WWLC2020FFAMOAIUBTOTOFV}]{%
\includegraphics[width=0.4925\linewidth]{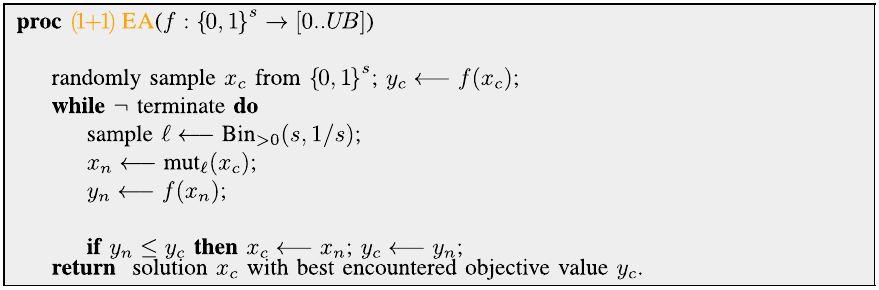}%
\label{fig:impl:ea}%
}%
\strut\hfill\strut%
\subfloat[\opofea~\cite{WWLC2020FFAMOAIUBTOTOFV}]{%
\includegraphics[width=0.4925\linewidth]{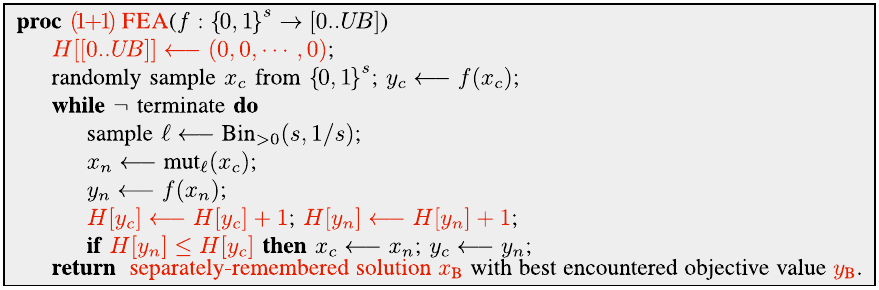}%
\label{fig:impl:fea}%
}%
\\\noindent%
\subfloat[\gga~\cite{CPD2017TAMPARAOEA,CPD2018ASPFTUOCIBBO} with $p=(1+\sqrt{5})/(2\scale)$]{%
\includegraphics[width=0.4925\linewidth]{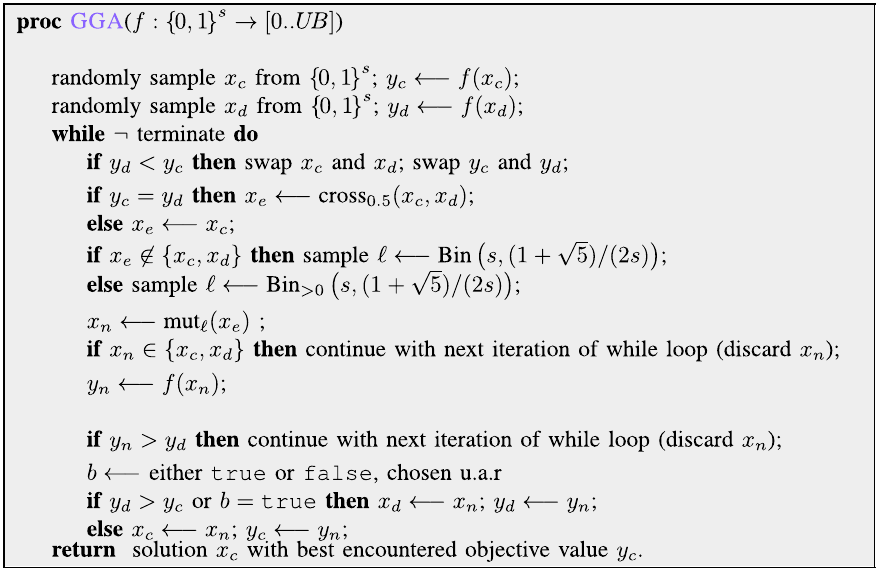}%
\label{fig:impl:gga}%
}%
\strut\hfill\strut%
\subfloat[\gfga\ with $p=(1+\sqrt{5})/(2\scale)$]{%
\includegraphics[width=0.4925\linewidth]{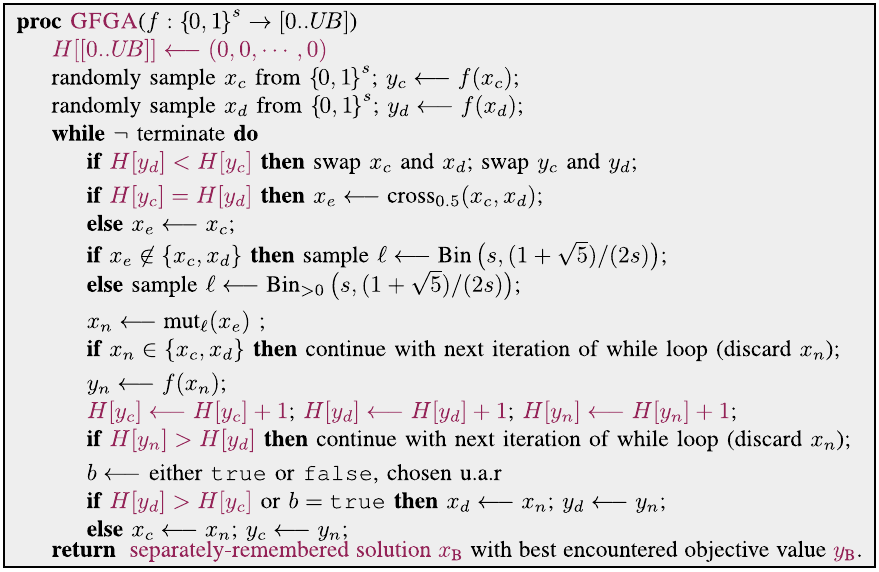}%
\label{fig:impl:gfga}%
}%
\\\noindent%
\subfloat[\saga~\cite{CPD2017TAMPARAOEA} with $F=1.5$, $p=\lambda/\scale$, $c=1/\lambda$]{%
\includegraphics[width=0.4925\linewidth]{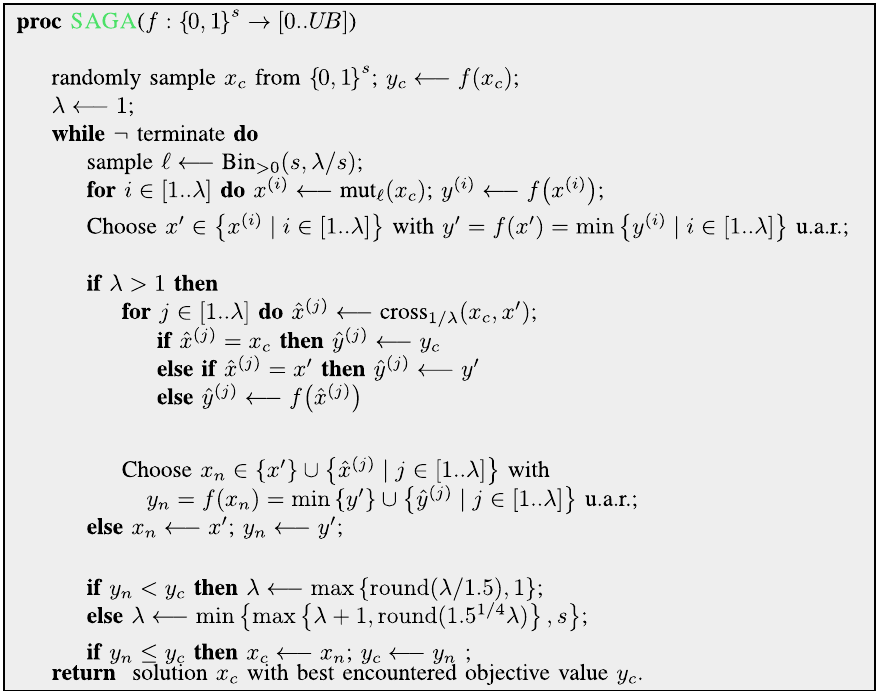}%
\label{fig:impl:saga}%
}%
\strut\hfill\strut%
\subfloat[\safga\ with $F=1.5$, $p=\lambda/\scale$, $c=1/\lambda$]{%
\includegraphics[width=0.4925\linewidth]{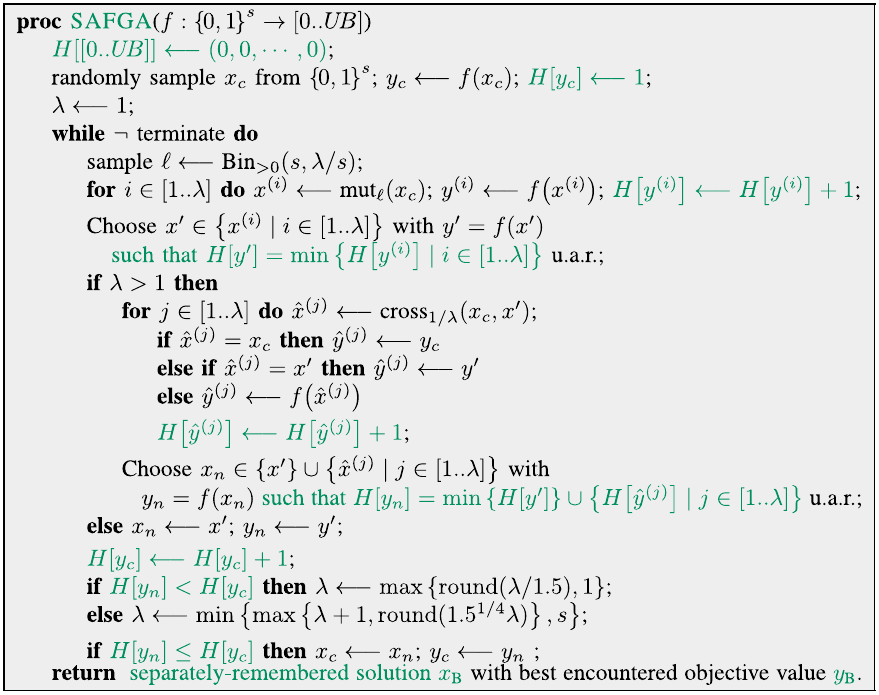}%
\label{fig:impl:safga}%
}%
\caption{%
The pseudo codes of six of the investigated algorithms for minimization. %
The left column contains the original variant, the right column the one with FFA, with differences highlighted. %
\emph{Note:}~The implementations of the FFA-based algorithm internally remember and return the candidate solution~\bestSoFarX\ with the best encountered \emph{objective value}~\bestSoFarF\ (not fitness).}%
\label{fig:algorithmsA}%
\end{figure*}

\label{sec:algorithms}%
Black-box metaheuristics working on bit strings~$\solutionSpace=\{0,1\}^{\scale}$ of a fixed length~\scale\ are a major concern for research.
We integrate FFA into the basic \opoeaP\ and two state-of-the-art theory-inspired GAs for minimization problems.
On the left-hand side of Figure~\ref{fig:algorithmsA}, we present the original algorithms as published in literature.
On the right-hand side, their corresponding variants using FFA are defined.

Without loss of generality, we limit our algorithm definitions to objective functions with a range~\intFromTo{0}{\upperBound} containing all integer numbers from~0 to~\upperBound.

We apply all algorithms in their modified efficient form defined in~\cite{CPD2017TAMPARAOEA} and also use the following notation from this work:
\mbox{$\ell\longleftarrow\text{Bin}(\scale, p)$} stands for sampling a number~$\ell$ from the binomial distribution with $\scale$~trials and success probability~$p$, i.e., \mbox{$P[\ell=k]={\scale \choose k}p^k(1-p)^k$}.
Setting \mbox{$\ell\longleftarrow\text{Bin}_{>0}(\scale,p)$} means to repeat this sampling until an $\ell>0$ is encountered.
The operation \mbox{$\text{mut}_{\ell}(\solspel)$} creates a copy of~$\solspel$ with exactly~$\ell$ bits flipped at different indexes chosen uniformly at random~\mbox{(u.a.r.).} 
\mbox{$\text{cross}_c(\solspeli{1},\solspeli{2})$} performs crossover by choosing, independently for every position, the value of the corresponding bit from~\solspeli{2} with probability~$c$ and from~\solspeli{1} otherwise.

The first algorithm in our study is the \opoeaP, which starts with a random bit string~\solspeli{c}.
In every step, it samples a new solution~\solspeli{n} as a modified copy of~\solspeli{c} by flipping each bit with probability~$1/\scale$.
This would correspond to drawing the number~$\ell$ of bits to flip from the binomial distribution with~$p=1/\scale$.
In the \opoea\ defined in Figure~\ref{fig:impl:ea}, we instead use \mbox{$\ell\longleftarrow\text{Bin}_{>0}(\scale,1/\scale)$} to ensure~$\ell>0$.
Thus, $\solspeli{n}\neq\solspeli{c}$ always holds and we avoid wasting function evaluations (FEs)~\cite{CPD2017TAMPARAOEA}.
If~\solspeli{n} is at least as good as~\solspeli{c}, i.e., $\objFunb{\solspeli{n}}\leq\objFunb{\solspeli{c}}$, then it replaces~\solspeli{c}.

We plugged FFA into this algorithm in~\cite{WWLC2020FFAMOAIUBTOTOFV} and obtained the \opofea\ (Figure~\ref{fig:impl:fea}).
The \opofeaP\ bases all decisions influencing the course of its search on the frequency fitness table~\ffaH.
It also internally maintains a variable \bestSoFarX\ in which it stores the solution with the best-ever encountered objective value~\bestSoFarF\ to be returned as end results.
We have omitted this detail for the sake of brevity in all FFA-based algorithms.

The \ggaL~\cite{S2012CSUBBA,S2017HCSUBBAIGA} retains a population of the best two solutions~\solspeli{c} and \solspeli{d} and without loss of generality, we assume that $\objFunb{\solspeli{c}}\leq\objFunb{\solspeli{d}}$.
In each iteration, it randomly chooses two parents with the best fitness in the population \emph{with replacement} to create an offspring~\solspeli{e} via uniform crossover~($c=0.5$).
It then flips each bit in~\solspeli{e} independently with probability~$p=(1+\sqrt{5})/(2\scale)$ to obtain the new solution~\solspeli{n}.
If~$\objFunb{\solspeli{n}}\leq\objFunb{\solspeli{d}}$, it will replace one of the parents: either~\solspeli{d}, if \solspeli{d}~is worse than~\solspeli{c}, otherwise either of the \mbox{two u.a.r.}

The \gga~\cite{CPD2017TAMPARAOEA,CPD2018ASPFTUOCIBBO} (Figure~\ref{fig:impl:gga}) ensures that crossover is only done if~$\objFunb{\solspeli{c}}=\objFunb{\solspeli{d}}$ and that the parents are chosen \emph{without replacement}, while otherwise $\solspeli{e}=\solspeli{c}$.
Furthermore, if $\solspeli{e}\in\left\{\solspeli{c},\solspeli{d}\right\}$, it is enforced that $\ell\geq 1$~bits are flipped in mutation.\footnote{%
We retain the mutation rate~$p$ from the original \ggaL\ algorithm~\cite{S2012CSUBBA,S2017HCSUBBAIGA}, which is not the optimal choice for our \gga\ algorithm variant on \onemax\gga~\cite{CPD2017TAMPARAOEA,CPD2018ASPFTUOCIBBO}.} 

The \ggaP\ was shown to be twice as fast as \emph{any} mutation-only EA on \onemax~\cite{S2012CSUBBA,S2017HCSUBBAIGA}.
We plug FFA into this algorithm and obtain the \gfga\ specified in Figure~\ref{fig:impl:gfga}, which allows us to investigate the impact FFA on an algorithm where crossover -- and hence, a (small) population -- are provably efficient.

The \ollga\ proposed in~\cite{DDE2015FBBCTDNGA} maintains a single best individual~\solspeli{c}.
In every step, it generates~$\lambda$ offspring~\solspels{i} using mutation probability~$p=k/\scale$ with~$k>1$.
The best of them, $\solspel'$, is then used as parent for another~$\lambda$ offspring~\solspelps{j} created via crossover using probability~$c$.
If~\solspeli{n}, the best among the offspring~\solspelps{j}, is at least as good as~\solspeli{c}, it replaces it.
The \sagaL~\cite{DDE2015FBBCTDNGA} builds on this algorithm by adjusting the parameter~$\lambda$:
If~\solspeli{n} is better than~\solspeli{c}, $\lambda$ is decreased to~$\lambda/F$ and otherwise increased to $F^{1/4}\lambda$.
$F=1.5$ and $\lambda$ always remains in~$\intFromTo{1}{\scale}$.
In~\cite{DD2015OPCTSAAT15TRIDS,DD2018OSASAPCFT1LLGA}, the parameters are set to~$p=\lambda/\scale$ and~$c=1/\lambda$.

The \saga~\cite{CPD2017TAMPARAOEA} in Figure~\ref{fig:impl:saga} improves the efficiency by ensuring that mutation flips $\ell\geq1$~bits, that crossover offspring equaling one of their parents are not evaluated, and by also letting~$\solspel'$ participate in the selection step after crossover.

The \sagaP\ is a theory-inspired GA with linear expected runtime on~\onemax~\cite{DD2018OSASAPCFT1LLGA}.
In Figure~\ref{fig:impl:safga}, we present the \safga, a variant of this algorithm applying FFA.
We can thus investigate the impact of the different optimization paradigm offered by FFA on this highly efficient algorithm and its process of self-adjustment, which now is based on the frequency fitness.

As a side note:
Not wasting FEs on repeatedly evaluating the same solutions saves runtime for the pure algorithm variants.
This does not necessarily hold for the FFA-based variants, as evaluating the same solution again would influence the frequency fitness and we could increment the frequency counters without actually invoking the objective function.
We do not investigate this approach here as it would require us to double the experimentation effort.
Still, we want to at least mention that the modifications from~\cite{CPD2017TAMPARAOEA} may not necessarily retain their impact under FFA.

\begin{figure}%
\centering%
\subfloat[\eafea]{%
\includegraphics[width=0.99\linewidth]{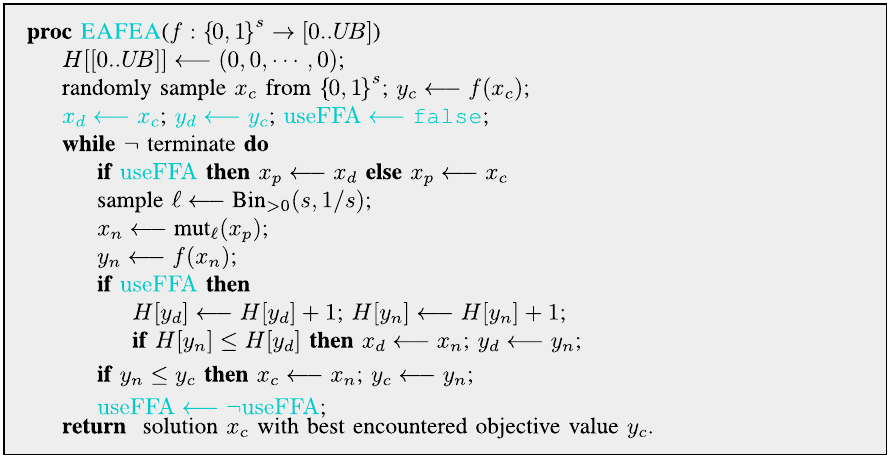}%
\label{fig:impl:eafea}%
}%
\\%
\subfloat[\safgap\ with $F=1.5$, $p=\lambda/\scale$, $c=1/\lambda$]{%
\includegraphics[width=0.99\linewidth]{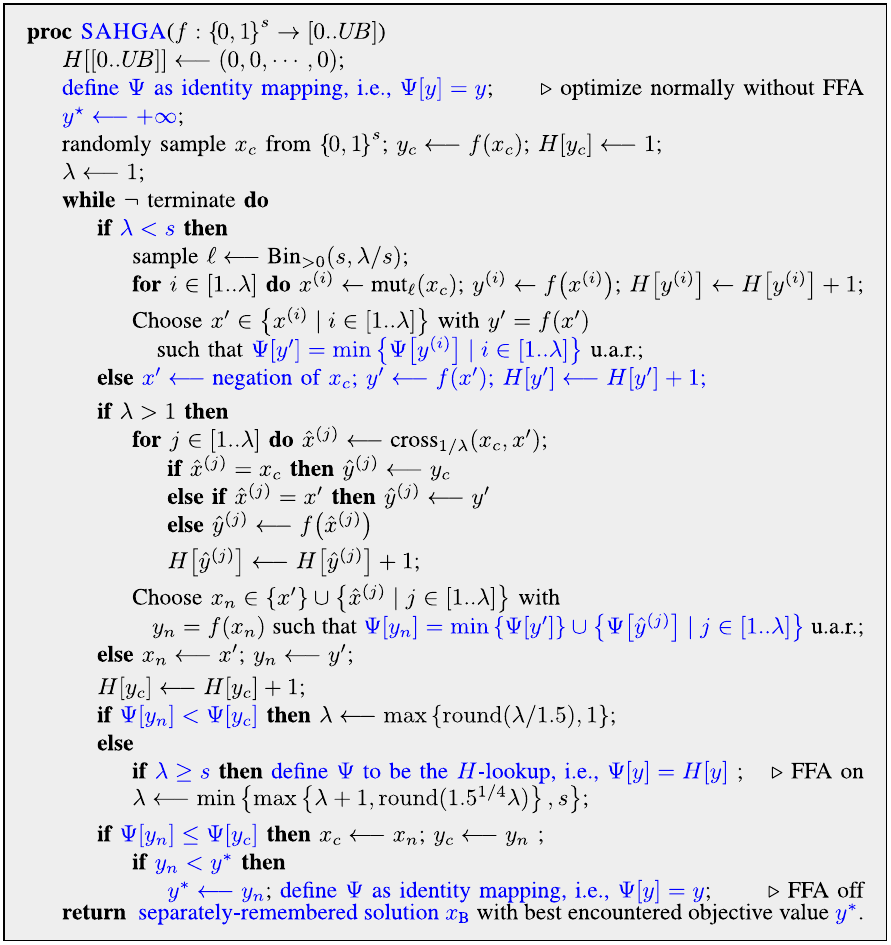}%
\label{fig:impl:safgap}%
}%
\caption{The hybrid algorithms using both FFA and direct optimization.}%
\label{fig:algorithmsB}%
\end{figure}%

Besides FFA-only and pure algorithm variants, we can also conceive hybrids.
One could, for instance, assign the FEs to the \opoeaP\ and the \opofeaP\ in a round robin fashion.
In this case, the resulting runtime on any problem would be at most twice the runtime of the faster of the two, making it uninteresting for experiments.
The \eafea\ in Figure~\ref{fig:impl:eafea} does this with a slight twist:
If the \opofeaP\ part discovers a solution equally good to the best-so-far solution of the \opoeaP\ part, it will overwrite it, thus ``informing'' the \opoeaP-part.

Another idea would be to toggle between FFA and pure optimization.
This concept is implemented in our \safgap\ in Figure~\ref{fig:impl:safgap}, which behaves exactly as the pure~\sagaP\ until one iteration after~$\lambda$ reaches~$\scale$, i.e., when basically all bits will be flipped during mutation.
At this moment, the algorithm switches over to using FFA and keeps using it until the best-so-far solution improves, at which point it will toggle back to pure optimization.
In \safgapP, we also added the minor improvement that only one single mutation is performed if $\lambda=\scale$, since then all bits are flipped and all mutation offspring are identical.

We thus cover two hybridization ideas, namely a round-robin like and a back-and-forth switching method.%
\section{Experiments}%
\label{sec:experiments}%
We now apply the eight algorithms to a wide selection of different benchmark problems.
The objective functions are all subject to minimization.
Their ranges are~$\objectiveSpace\subseteq\intFromTo{0}{\upperBound}$, where the upper bounds~\upperBound\ are either linear functions of the problem scale~\scale\ or a small polynomial of it.
On each problem instance and for each algorithm, except for the \maxsat\ instances, we conduct at least 71~independent runs.
Except for \onemax, \leadingones, and \nqueens, we limit all runs to at most $10^{10}$~FEs in our experiments.%
\begin{figure}[tb]%
\centering%
\includegraphics[width=0.99\linewidth]{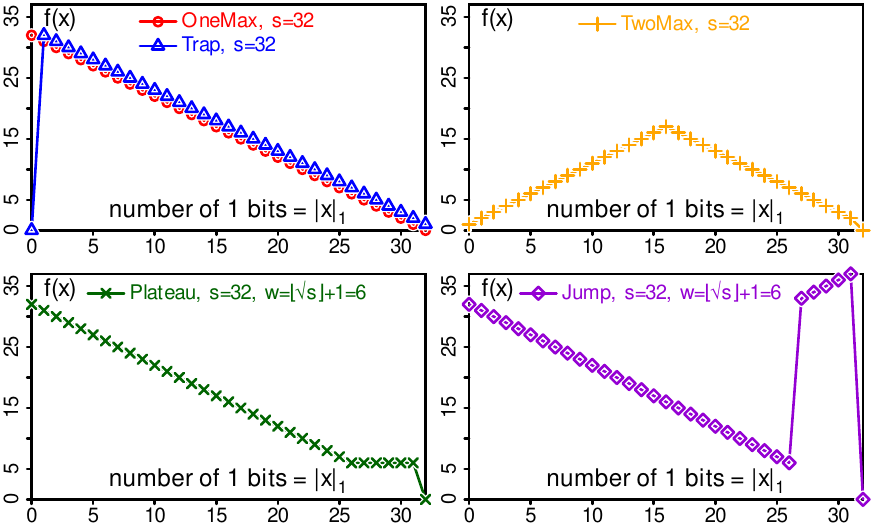}
\caption{Illustrations of the \onemax, \twomax, \trap, \jump, and \plateau\ problems for $\scale=32$ and $\jumpWidth=6$.}%
\label{fig:function_illustrations}%
\end{figure}%
\subsection{\onemax~Problem}%
\label{sec:onemax}%
\begin{figure}[tb]%
\centering%
\includegraphics[width=0.99\linewidth]{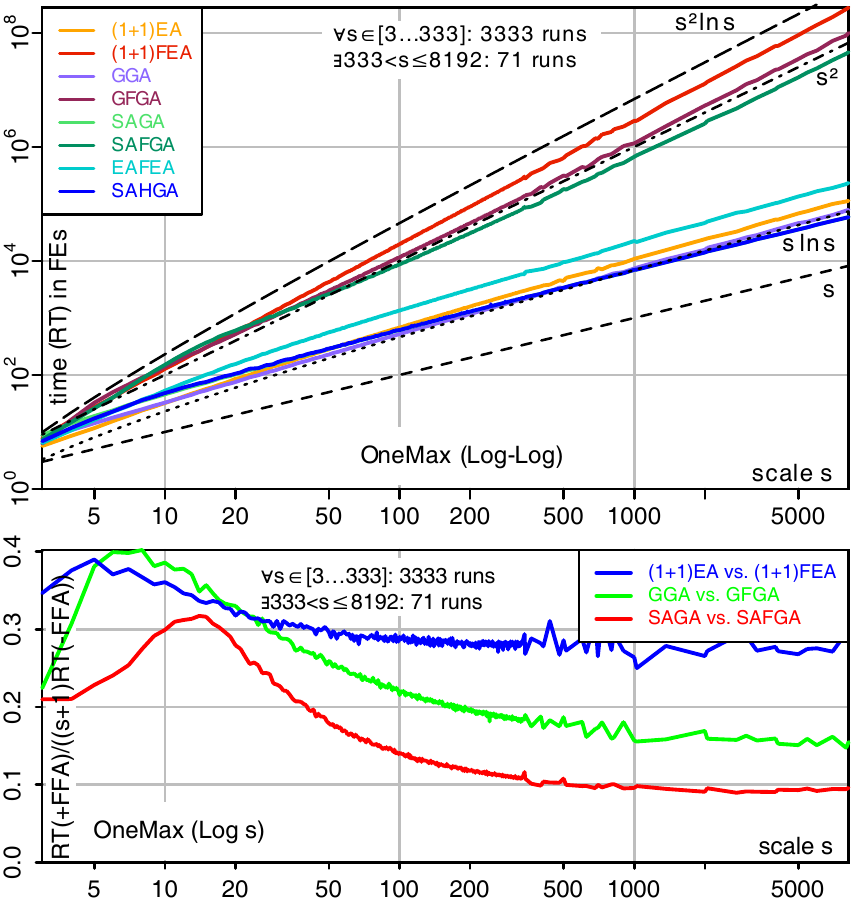}
\caption{The mean runtimes measured on selected problem sizes~\scale\ (horizontal axes) of the \onemax\ problem. Top: The vertical axis shows the mean runtime. Bottom: The vertical axis shows the mean runtimes of the algorithms with FFA divided by $\scale+1$~times the mean runtime of their pure variants.}%
\label{fig:onemax_runtime}%
\end{figure}%
\onemax~\cite{M1992HGARWMAH} is a unimodal optimization problem where the goal is to discover a bit string of all ones.
Its minimization version of scale~\scale\ is defined below and illustrated in Figure~\ref{fig:function_illustrations}:%
\begin{equation}%
\onemax(\solspel) = \scale - \countones{\solspel}\textnormal{~where~}\countones{\solspel}=\sum_{i=1}^{\scale} \solspelval{i}%
\end{equation}%
\onemax\ has a black-box complexity of $\bigOmegaOf{\scale/\ln{\scale}}$~\cite{DJW2006UALBFRSHIBBO}.
The \opoeaP~\cite{M1992HGARWMAH,DDE2015FBBCTDNGA} and the faster \ggaP~\cite{S2012CSUBBA,S2017HCSUBBAIGA,DDE2015FBBCTDNGA} have an expected runtime in~$\bigThetaOf{\scale\ln{\scale}}$~FEs, while the \sagaP\ achieves~$\bigOof{\scale}$~\cite{DD2015OPCTSAAT15TRIDS,DD2018OSASAPCFT1LLGA}.
This is visible in the upper half of Figure~\ref{fig:onemax_runtime}, which also shows that the FFA-based variants of the algorithms are consistently slower.
The \eafeaP\ needs about twice the time of the \opoeaP.
The \safgapP\ performs identical to the \sagaP, such that the curve of the former exactly covers the curve of the latter, meaning that the switch condition to FFA is never reached.

In~\cite{WWLC2020FFAMOAIUBTOTOFV}, we speculated that the \opofeaP\ is slower than the \opoeaP\ by a factor linear in the number of different possible objective values, which is~\mbox{$\scale+1$} in case of \onemax\ as~\mbox{$\objectiveSpace=\intFromTo{0}{\scale}$}.
Such a linear relationship seems to hold for all three algorithms, as the lower part of Figure~\ref{fig:onemax_runtime} reveals a slowdown approaching algorithm-specific constants \mbox{$\zeta_1\in(0.05,0.45)$} times~\mbox{$s+1$}.

The \onemax\ problem is also suitable to check whether our implementations of the basic algorithms are correct.
In Figure~3 of~\cite{CPD2017TAMPARAOEA}, the \sagaP\ has a mean runtime right in the middle between 30'000 and 40'000~FEs to solve \onemax\ with $\scale=5000$, while our implementation needs approximately 35'590~FEs.
The \ggaP\ in~\cite{CPD2017TAMPARAOEA} has a mean runtime of slightly above 38'000~FE on this problem.
It uses a different parameter setting for the mutation probability ($p=0.773581/\scale$) than our implementation ($p=(1+\sqrt{5})/(2\scale)$).
If we use the setting from~\cite{CPD2017TAMPARAOEA}, we obtain a mean runtime of 38'502~FEs.
Both results replicate the performance observed in~\cite{CPD2017TAMPARAOEA} and we can be confident that our implementations are correct.%
\subsection{\leadingones~Problem}%
\begin{figure}[tb]%
\centering%
\includegraphics[width=0.99\linewidth]{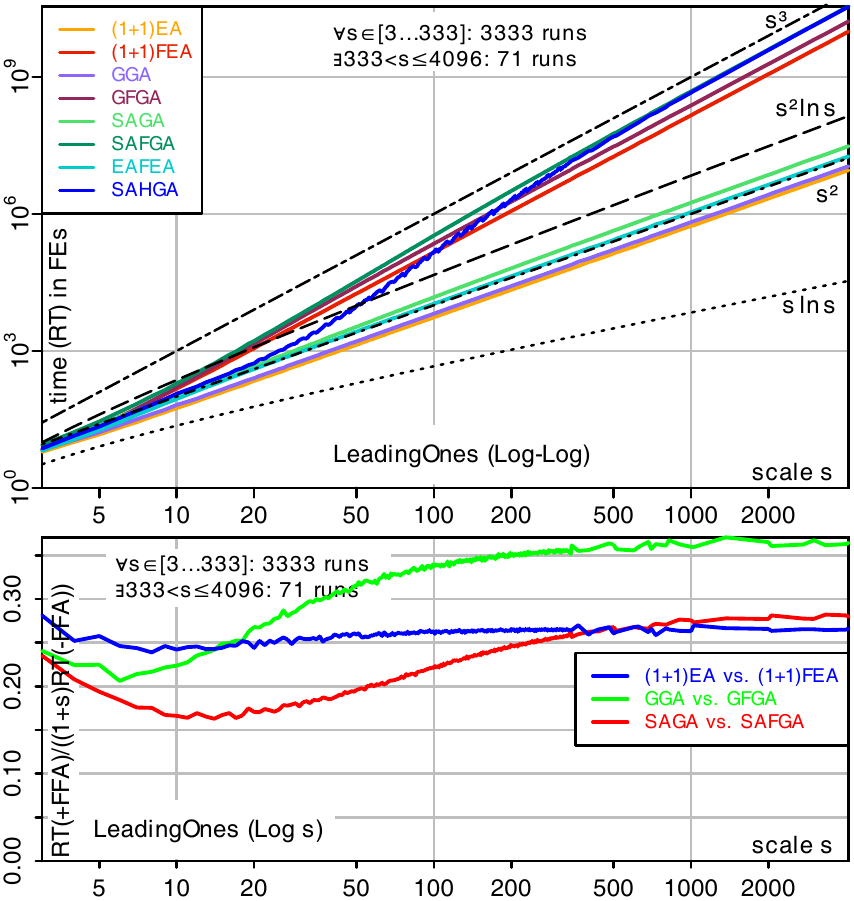}
\caption{The mean runtimes measured on selected problem sizes~\scale\ (horizontal axes) of the \leadingones\ problem. Explanation: See Figure~\ref{fig:onemax_runtime}.}%
\label{fig:leadingones_runtime}%
\end{figure}%
The \leadingones\ problem~\cite{W1989TGAASPWRBAORTIB,R1997CPOEA} maximizes the length of a leading sequence containing only 1~bits.
Its minimization version of scale~\scale\ is defined as follows:%
\begin{equation}%
\leadingones(\solspel) = \scale - \sum_{i=1}^{\scale} \prod_{j=1}^i \solspelval{j}%
\end{equation}%
The black-box complexity of \leadingones\ is~$\bigThetaOf{\scale\ln{\ln{\scale}}}$~\cite{AADLMW2013TQCOFAHP,AADDLM2019TQCOAPBVOM}.
The \opoeaP\ has a quadratic expected runtime on \leadingones~\cite{DJW2002OTAOTOPOEA} and so does the \ollga\ regardless of the value of~$\lambda$~\cite{ADK2019ATRAFTOPLCLGOL,KAD2019TAESOTOPLCLEOTLP}.
From Figure~\ref{fig:leadingones_runtime}, we find that the \sagaP\ performs worse on \leadingones\ than the \ggaP\ and \opoeaP, which behave similar.
The \eafeaP\ again needs about twice the time as the \opoeaP.
\safgapP\ initially performs similar to \sagaP, but as the scale~\scale\ increases, its curve approaches the one of the slower~\safgaP.
This means that the \sagaP\ reaches the switch condition to FFA on \leadingones, which explains its rather poor performance on this problem.
The FFA variants are again slower and again the lower part of the figure reveals a slowdown seemingly approaching constants $\zeta_2\in(0.1,0.4)$ times the number of different objective values.

The mean runtime of~14'980~FEs of our \sagaP\ implementation over 3333~runs on \leadingones\ with $\scale=100$ also fits well to the 15'920~FEs measured over 11~runs in the \myhref{https://github.com/IOHprofiler/IOHdata}{\texttt{IOHprofiler}} data set~\cite{DYHWSB2020BDOHWI}.%
\subsection{\twomax~Problem}%
\begin{figure}[tb]%
\centering%
\includegraphics[width=0.99\linewidth]{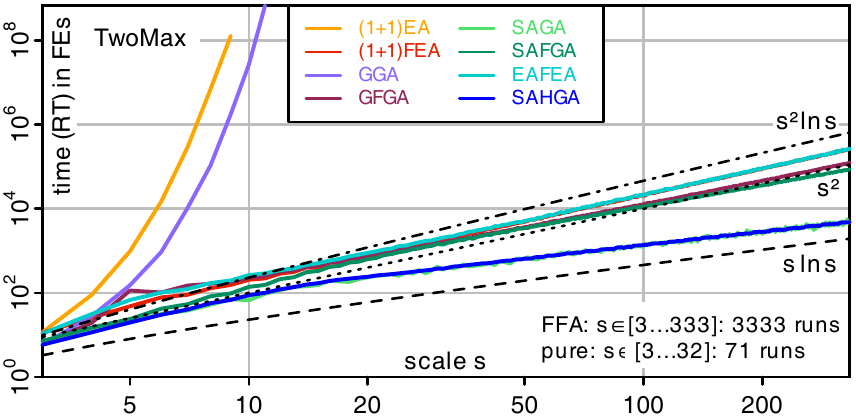}
\caption{The mean runtime measured on the \twomax\ problem.}%
\label{fig:twomax_runtime}%
\end{figure}%
The minimization version of the \twomax~\cite{FQW2018ELDBOAWHTMO,VHGN2002FTTIMEAHSP} problem of scale~\scale\ can be defined as follows:%
\begin{equation}%
\resizebox{0.905\linewidth}{!}{\ensuremath{%
\twomax(\solspel) = \left\{\!\!\!%
\begin{array}{r@{~}l}%
0&\textnormal{if }\countones{\solspel} = \scale\\%
1+\scale-\max\{\countones{\solspel}, \scale-\countones{\solspel}\}&\textnormal{otherwise}%
\end{array}%
\right.%
}}%
\end{equation}%
The \twomax\ problem introduces deceptiveness in the objective function by having a local and a global optimum.
Since their basins of attraction have the same size, a \opoeaP\ can solve the problem in \bigThetaOf{\scale\ln{\scale}} steps with probability~0.5 while otherwise needing exponential runtime in expectation, leading to a total expected runtime in~\bigOmegaOf{\scale^{\scale}}~\cite{FOSW2009AODPMFGE,FQW2018ELDBOAWHTMO}.
From Figure~\ref{fig:twomax_runtime}, we find that the \ggaP\ seems to behave similarly.
While it is a bit faster, it, too, can only solve small-scale instances of \twomax\ within the prescribed $10^{10}$~FE budget.
The \sagaP\ and the identically performing \safgapP\ do not suffer from the exponential runtime and solve any scale of the problem to which they were applied relatively quickly.
The reason is that if these algorithms arrive at the local optimum, they will not improve.
$\lambda$~will increase until reaching~$\scale$, at which point mutation will toggle all bits and jump directly to the global optimum.
From this explanation, we can expect that the \sagaP\ would still require exponential runtime if the two optima would not be bit-wise inverse of each other.
\eafeaP\ performs almost exactly like \opofeaP.

\opofeaP, \gfgaP, and \eafeaP\ can solve the \twomax\ instances for all $\scale\leq333$ within a mean runtime of below $\scale^2\ln{\scale}$~FEs.
\safgaP\ is faster than the above mentioned algorithms, but slower than \sagaP.
Since \sagaP\ needs more than $\scale\ln{\scale}$~FEs to solve the problem in average while \safgaP\ needs less than $\scale^2$, the experimentally observed slowdown seems to be less than linear this time.

From this moment on, we will often observe the following pattern regarding the FFA-based algorithm variants:
If the basic algorithm cannot efficiently solve the problem, its FFA-based variant can.
If the basic algorithm can solve the problem well, the FFA-based variant can still solve it but it is slower.%
\subsection{\trap\ Function}%
\begin{figure}[tb]%
\centering%
\includegraphics[width=0.99\linewidth]{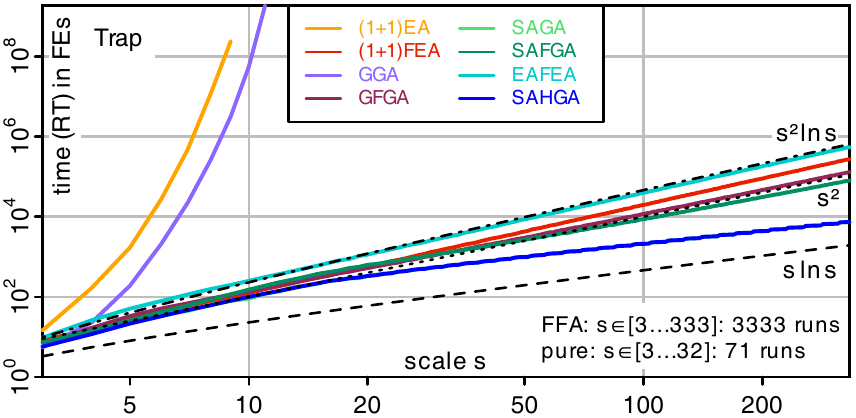}%
\caption{The mean runtimes measured on the \trap\ problem.}%
\label{fig:trap_runtime}%
\end{figure}%
The \trap\ function~\cite{NB2003AAOTBOSEAOTF,DJW2002OTAOTOPOEA} swaps the worst possible solution with the global optimum.
The \opoeaP\ here has an expected runtime of~\bigThetaOf{\scale^{\scale}}~\cite{DJW2002OTAOTOPOEA}.
The minimization version of the \trap\ function can be specified as follows:%
\begin{equation}%
\trap(\solspel) = \left\{\!\!\!%
\begin{array}{r@{~}l}%
0&\textnormal{if~}\countones{\solspel}=0\\%
\scale-\countones{\solspel}+1&\textnormal{otherwise}%
\end{array}\right.%
\end{equation}%
The \trap\ function is bijective transformation of the \onemax\ problem.
This means that all FFA-based algorithms behave exactly as on \onemax\ and can efficiently solve it.
\sagaP\ and \safgapP\ again perform the same, and for the same reason as on \twomax.
Figure~\ref{fig:trap_runtime} shows that all algorithms also behave similar as on the \twomax\ problem.%
\subsection{\jump~Problems}%
The \jump\ functions~\cite{DJW2002OTAOTOPOEA,FQW2018ELDBOAWHTMO,QGWF2021EAASFBOHTM} introduce a deceptive region of width~\jumpWidth\ with very bad objective values right before the global optimum.
The minimization version of the \jump\ function of scale~\scale\ and jump width~\jumpWidth\ is defined as follows:%
\footnote{%
Researchers have formulated different types of \jump\ functions. %
The one in~\cite{DDK2015UBBCOJF}, e.g., is similar to our \plateau\ function but differs in the plateau objective value.%
}%
\begin{equation}%
\jump(\solspel) = \left\{\!\!\!%
\begin{array}{r@{~}l}%
\scale-\countones{\solspel}&\textnormal{if }\left(\countones{\solspel} = \scale\right)\lor\left(\countones{\solspel} \leq \scale-\jumpWidth\right)\\%
\jumpWidth+\countones{\solspel}&\textnormal{otherwise}%
\end{array}%
\right.%
\label{eq:jump}%
\end{equation}%%
The expected runtime of the \opoeaP\ on such problems is in \bigThetaOf{\scale^{\jumpWidth}+\scale\ln{\scale}}~\cite{DJW2002OTAOTOPOEA}.%
\begin{figure}[tb]%
\centering%
\includegraphics[width=0.99\linewidth]{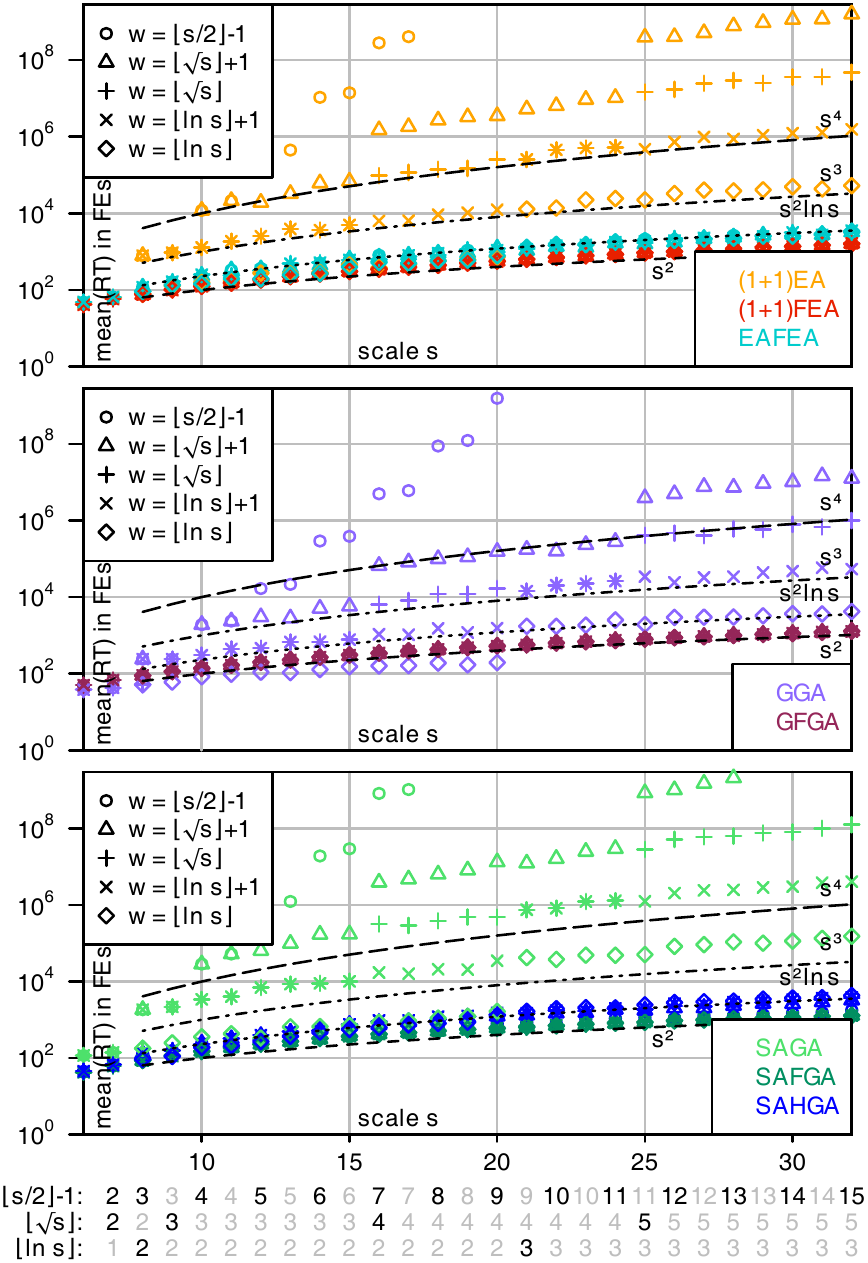}%
\caption{The mean runtime measured on \jump\ problems with scale~\scale\ and jump width~$\jumpWidth>1$.}%
\label{fig:jump_runtime}%
\end{figure}

We conduct experiments with five different jump widths~\jumpWidth, namely %
\mbox{$\left\lfloor\ln{\scale}\right\rfloor$}, %
\mbox{$\left\lfloor\ln{\scale}\right\rfloor+1$}, %
\mbox{$\left\lfloor\sqrt{\scale}\right\rfloor$}, %
\mbox{$\left\lfloor\sqrt{\scale}\right\rfloor+1$}, and %
\mbox{$\left\lfloor0.5\scale\right\rfloor-1$}. %
From Figure~\ref{fig:jump_runtime}, we find that the performance of all pure algorithms quickly deteriorates with rising~\jumpWidth.
Since the \jump\ problem is another bijective transformation of the \onemax\ problem, all FFA-based algorithms behave the same as on \onemax.
Indeed, \jump, \trap, and \onemax\ all are identical from the perspective of an algorithm that utilizes FFA, which is one of the properties making this concept interesting.
\safgapP\ and \eafeaP\ are slightly slower, but, too, can solve the \jump\ problem efficiently regardless of the jump width.%
\subsection{\plateau~Problems}%%
\begin{figure}[tb]%
\centering%
\includegraphics[width=0.99\linewidth]{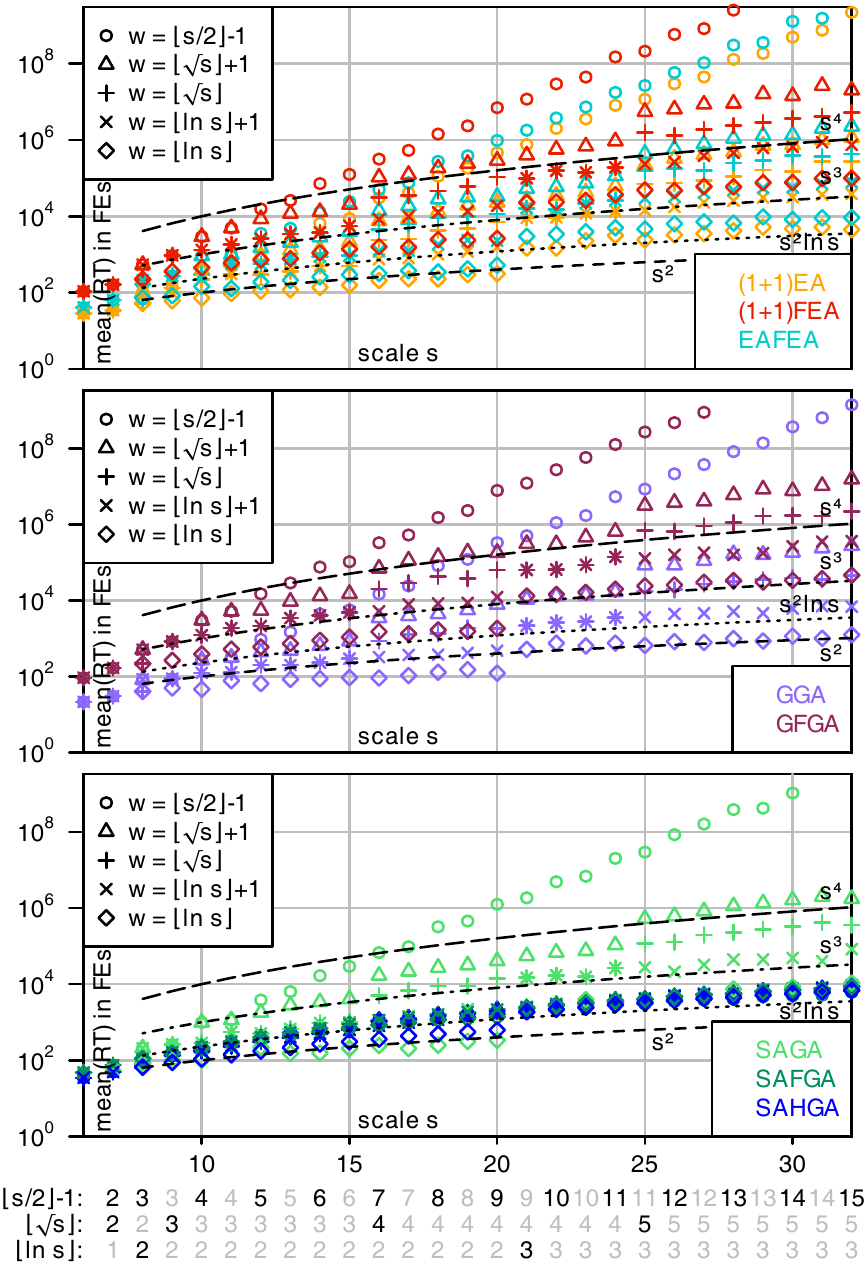}%
\caption{The mean runtime measured on \plateau\ problems with scale~\scale\ and plateau width~$\plateauWidth>1$.}%
\label{fig:plateau_runtime}%
\end{figure}%
\begin{figure}[tb]%
\centering%
\includegraphics[width=0.99\linewidth]{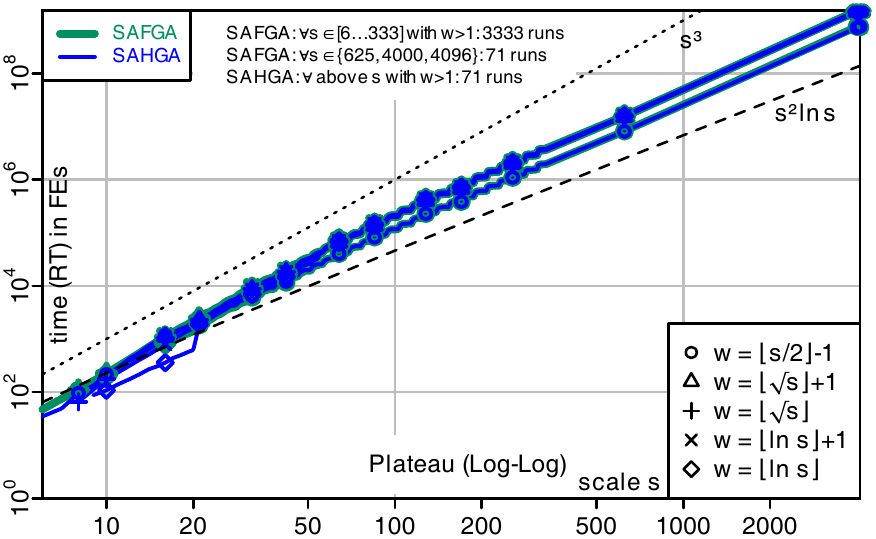}%
\caption{A more comprehensive experiment with \safga\ and \safgap\ on the \plateau\ problem.}%
\label{fig:plateau_saga_runtime}%
\end{figure}%
The minimization version of the \plateau~\cite{AD2021PRAFPF} function of scale~\scale\ with plateau width~\plateauWidth\ is defined as follows:%
\begin{equation}%
\resizebox{0.905\linewidth}{!}{\ensuremath{%
\plateau(\solspel) = \left\{\!\!\!%
\begin{array}{r@{~}l}%
\scale-\countones{\solspel}&\textnormal{if }\left(\countones{\solspel} = \scale\right)\lor\left(\countones{\solspel} \leq \scale-\plateauWidth\right)\\%
\plateauWidth&\textnormal{otherwise}%
\end{array}%
\right.%
}}%
\end{equation}%
The expected runtime of the \opoeaP\ on such a problem is in~\bigThetaOf{\scale^{\plateauWidth}}~\cite{AD2021PRAFPF}.
Figure~\ref{fig:plateau_runtime} shows that, while the pure algorithms can solve the \plateau\ problem better than the \jump\ problem, they are slow at doing so.

A plateau of the objective function is also a plateau under FFA.
Indeed, we can observe that here, FFA provides no advantage for the \opofeaP\ and \gfgaP, which are slower than the \opoeaP\ and \ggaP, respectively.
The \safgaP\ and the \safgapP\ seemingly perform almost the same on all \plateau\ instances of a given scale~\scale\ in Figure~\ref{fig:plateau_runtime}, regardless of the plateau width~\plateauWidth, \emph{and} solve them efficiently (and better than \sagaP).

We repeat the experiment with both algorithms for larger scales up to~4096 in Figure~\ref{fig:plateau_saga_runtime}.
We find that \safgaP\ and \safgapP\ solve the problem consistently in less than $\scale^4$~FEs.
From the figure, it seems that the large plateaus with \mbox{$\plateauWidth=\left\lfloor0.5\scale\right\rfloor-1$} are \emph{easier} for both algorithms and the other four plateau widths lead to similar performance.
When comparing the raw numbers, we find that problem tends to get easier for the algorithms the wider the plateaus are, i.e., the bigger~$\plateauWidth$.\footnote{%
The difference between \mbox{$\left\lfloor\ln{\scale}\right\rfloor$} and \mbox{$\left\lfloor\sqrt{\scale}\right\rfloor+1$} is just too small for the investigated scales for this to be visible in the diagram.}

This can be explained as follows:
As shown in~\cite{WWLC2020FFAMOAIUBTOTOFV}, FFA-based algorithms may optimize either towards improving solutions or towards worsening solution quality and from time to time change direction.
If the algorithm switches towards worsening objective values after reaching the plateaus, it will eventually reach the worst-possible bit string~$\solspel_w$, composed of all~\texttt{0}s.
There is only one such string and all strings with some~\texttt{1}s in them may have thus already received higher frequency values.
Success in the FFA-based variants corresponds to finding a less-frequently encountered objective value and a failure in doing so increases~$\lambda$.
Once~$\solspel_w$ is reached, the algorithms may not immediately succeed to find a string with lower objective value encounter frequency.
The self-adjustment will thus increase~$\lambda$, leading to more bits being flipped at once in mutation.
This, in turn, increases the chance to directly jump into the plateau (which may already have a high frequency assigned to it) and does so the more, the larger~$\plateauWidth$.
Once $\lambda=\scale$ is reached, all bits will be flipped, jumping from~$\solspel_w$ directly to the global optimum.
Based on this explanation, we can also expect that a problem where plateaus surround both the optimum and~$\solspel_w$ would still be hard for the \safgaP\ and \safgapP.%
\subsection{\nqueens\ Problem}%
\begin{figure}[tb]%
\centering%
\includegraphics[width=0.99\linewidth]{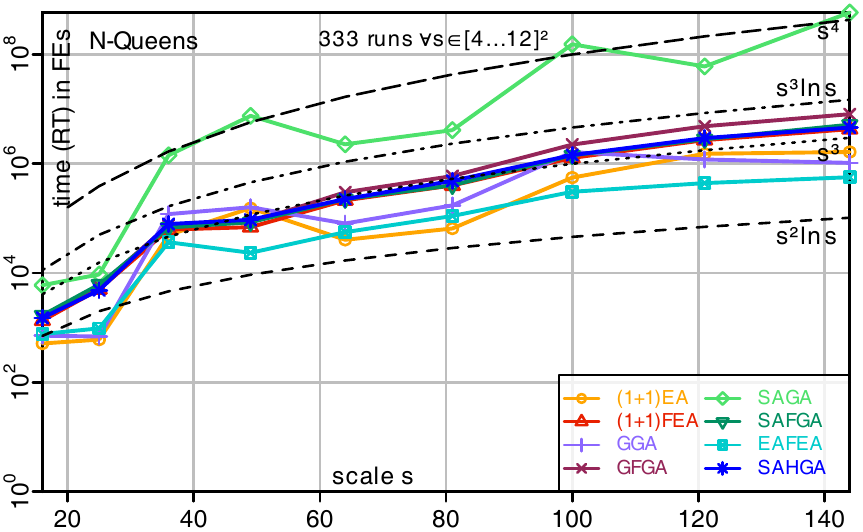}
\caption{The mean runtime measured on the \nqueens\ problem.}%
\label{fig:nqueens_runtime}%
\end{figure}%
The \nqueens\ problem is defined for bit strings of length~$\scale=\nqueensN^2$~\cite{DYHWSB2020BDOHWI}.
A bit string~\solspel\ is mapped to a chess board and a queen is placed for any bit of value~1.
The goal is to place \nqueensN~queens such that they cannot attack each other.
The total number of queens on the board be \mbox{$Q(\solspel)= \countones{\solspel}$}, which might be more or less than~\nqueensN.
We also count the \mbox{number $Q_{\xi}(\solspel)$} of queens in every single row, column, and \mbox{diagonal~$\xi$} of the chess board. 
The minimization version of the \nqueens\ problem is then:%
\begin{equation}%
\nqueens(\solspel)=\nqueensN-Q(\solspel) + \nqueensN\sum_{\forall\xi} \max\{0, Q_{\xi}(\solspel)-1\}%
\end{equation}%
The \nqueens\ problems are attested moderate difficulty in~\cite{DYHWSB2020BDOHWI}.
They are interesting for investigating FFA because their number of possible objective values is larger than their scale~\scale.

In this experiment, \sagaP\ shows somewhat unstable performance.
We therefore investigate all $\nqueensN\in\intFromTo{4}{12}$ without runtime limit and 333~independent runs per algorithm and problem instance. 
For~$\nqueensN=11$, one of these \sagaP\ runs needed 11'936'590'163~FEs and for~$\nqueensN=12$, \sagaP\ needed 120'878'587'950~FEs in one run. This could indicate a bad, potentially exponential, worst-case runtime.
The next-worst longest runtime over the 333~runs of any other algorithm is by the \opoeaP\ for $\nqueensN=12$ with 336'432'114~FEs.

Figure~\ref{fig:nqueens_runtime} confirms that the \sagaP\ performed worst.
Its FFA-based variants perform much better (both need about 28'000'000~FEs for $\nqueensN=12$ in their worst runs) and similar to the \opofeaP.
The \eafeaP\ performed best for larger scales, followed by \ggaP\ and \opoeaP.%
\subsection{\isingod\ Problem}%
\begin{figure}[tb]%
\centering%
\includegraphics[width=0.99\linewidth]{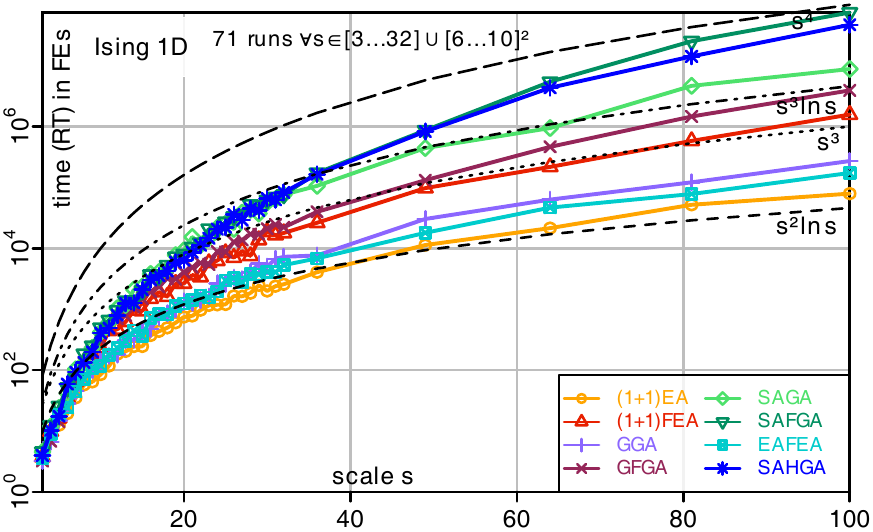}
\caption{The mean runtime measured on the \isingod\ problem.}%
\label{fig:ising1d_runtime}%
\end{figure}%
Ising problems define a graph~$G(V=\intFromTo{0}{\scale-1}, E\subseteq V^2)$ where each node stands for one bit index and the edges are undirected.
If two indices~$i$ and~$j$ are connected, a penalty of~1 is incurred in the objective value of a solution~\solspel\ if $\solspel[i]\neq\solspel[j]$, as shown in Equation~\eqref{eq:ising}.
In \isingod\ problem~\cite{FW2005TODIMMVR,DYHWSB2020BDOHWI}, $G$ describes a one-dimensional ring where each bit is connected to its predecessor and successor in~\solspel\ and $E=E_1$ is then defined in Equation~\eqref{eq:isingE1d}.%
\begin{eqnarray}%
\ising(\solspel,E)=\sum_{\forall(i,j) \in E} \left|\solspel[i]-\solspel[j]\right|%
\label{eq:ising}%
\\%
E_1=\left\{(i,j)\mid i,j\in\intFromTo{0}{\scale-1}\land j=(i+1)\bmod\scale\right\}%
\label{eq:isingE1d}%
\end{eqnarray}%
From \cite{FW2005TODIMMVR} we know that the \opoeaP\ has an expected running time of~\bigOof{\scale^3} on this problem with small constant in the \bigO-term, which is visible in Figure~\ref{fig:ising1d_runtime}, where its observed mean running time is between~\mbox{$\scale^2\ln{\scale}$} and~$\scale^3$.
The simple \opoeaP\ beats \ggaP, which, in turn, beats \sagaP.
We further find that all FFA-based algorithm variants are slower than their corresponding pure variants, but again by no more than a factor linear in~\scale.
All algorithms solved all instances in all runs within the prescribed budget.%
\subsection{\isingtd\ Problem}%
\begin{figure}[tb]%
\centering%
\includegraphics[width=0.99\linewidth]{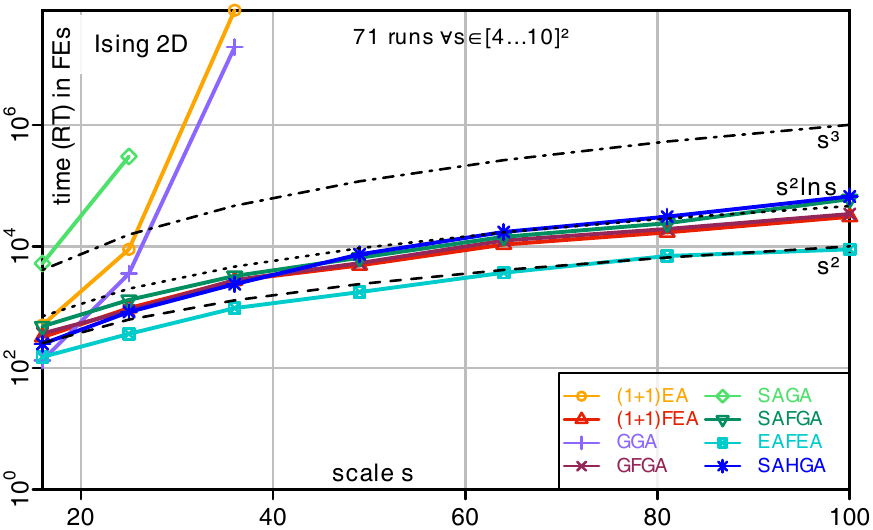}
\caption{The mean runtime measured on the \isingtd\ problem.}%
\label{fig:ising2d_runtime}%
\end{figure}%
In the \isingtd\ problem~\cite{F2004APUBFAMBAOTTDIM,DYHWSB2020BDOHWI}, the graph plugged into Equation~\eqref{eq:ising} describes a two-dimensional torus with~$\scale=N^2$ and~$E=E_2$, where each bit has four neighbors:%
\begin{equation}%
E_2=\begin{array}[t]{l}%
\{ (\alpha+\beta N,\gamma+\delta N)\;\mid\;%
\alpha,\beta,\gamma,\delta\in\intFromTo{0}{N-1}\;\mathrlap{\land}\\\relax%
\quad[(\gamma=(\alpha+1)\bmod N \;\land \;\delta=\beta)\;\lor\\\relax%
\quad\phantom{[}(\gamma=\alpha \;\land \;\delta=(\beta+1)\bmod N)]\}%
\end{array}%
\end{equation}%
We investigate the \isingtd\ problem up to $N=10$, i.e., $\scale=100$.
Figure~\ref{fig:ising2d_runtime} reveals that this problem is much harder than \isingod\ for the pure algorithms, which can only reach a 100\% success rate until~$N=6$ within the $10^{10}$~FEs budget.
This fits to~\cite{F2004APUBFAMBAOTTDIM}, which states that the \opoeaP\ has exponential expected running times on a similar class of two-dimensional Ising models.
All variants with FFA, however, can solve all \isingtd\ instances in all of their runs.
Oddly, for them \isingtd\ seems to be \emph{easier} than \isingod\ for the \opoeaP, which was the fastest algorithm on it!
The \eafeaP\ performed best by a margin.
We confirmed the correct implementation of both problems with unit tests.%
\subsection{Linear Function with Harmonic Weights}%
\begin{figure}[tb]%
\centering%
\includegraphics[width=0.99\linewidth]{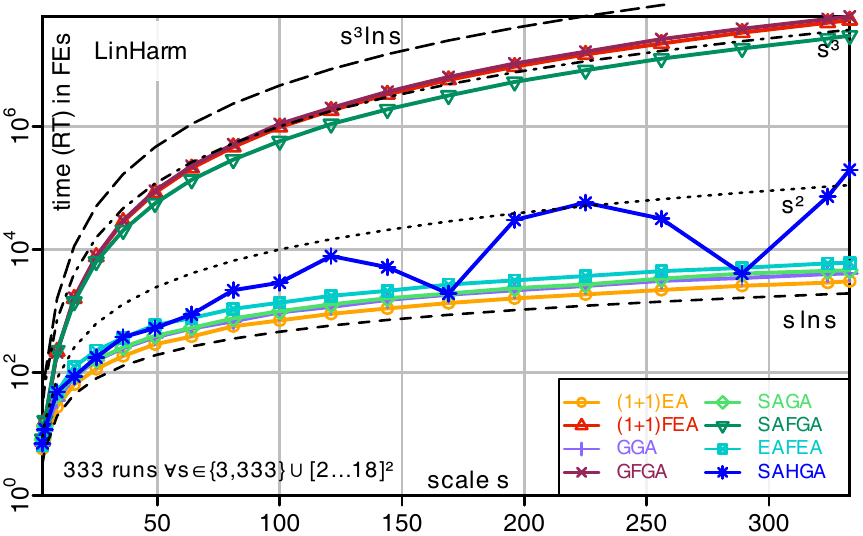}%
\caption{The mean runtime measured on the \linearHarmonic\ problem.}%
\label{fig:linear_harmonic_runtime}%
\end{figure}%
The minimization version of the linear harmonic function~\cite{DYHWSB2020BDOHWI} is given in Equation~\eqref{equ:linharm}.
The \opoeaP\ can solve this function in~\bigThetaOf{\scale\log\scale}~FEs~\cite{DYHWSB2020BDOHWI,DJW2002OTAOTOPOEA}.%
\begin{equation}%
\linearHarmonic(\solspel) = 0.5\scale(\scale+1) - \sum_{i=1}^{\scale} i\,\solspeli{i}%
\label{equ:linharm}%
\end{equation}%
This problem is particularly interesting for investigating FFA, because it is rather easy to solve, but its number of objective values grows quadratically with~\scale.
From Figure~\ref{fig:linear_harmonic_runtime}, we see that the FFA-only variants require mean runtimes which could indeed be quadratically larger than the mean runtimes of the respective pure algorithms.
This indicates that the cost of FFA on easy problems is likely proportional to the number of possible objective values.%
\subsection{\wmodel\ Instances}%
\begin{figure}[tb]%
\centering%
\includegraphics[width=0.99\linewidth]{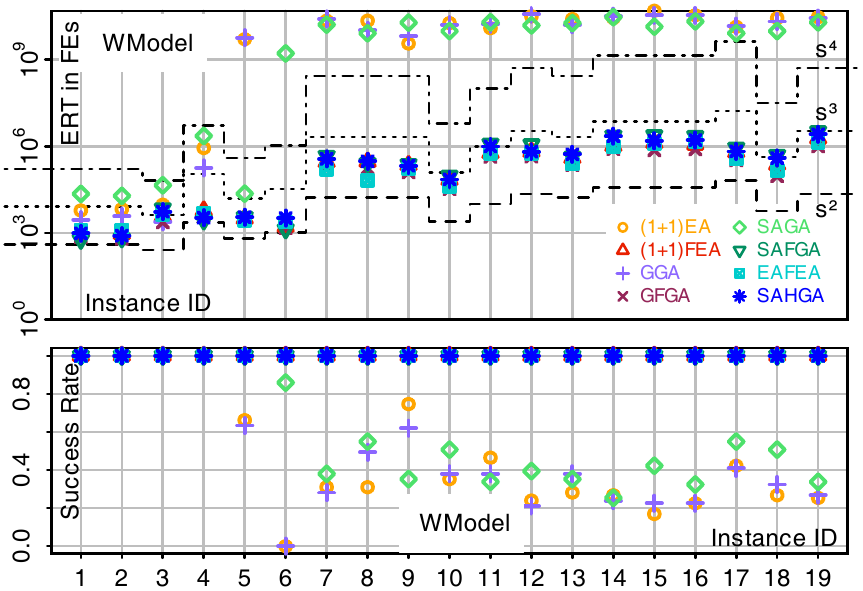}%
\caption{The \ert\ (top) and success rates (bottom) measured on the 19~\wmodel\ instances from~\cite{WCLW2019SADSOBIFATMPFBBDOA}.}%
\label{fig:wmodel_results}%
\end{figure}%
The \wmodel~\cite{WW2018DFOCOPATTWMBPFST,WNSRG2008ATMFMOERANFL} is a benchmark problem which exhibits different difficult fitness landscape features (ruggedness, deceptiveness, neutrality, and epistasis) in a tunable fashion.
19~diverse \wmodel\ instances of different \mbox{scales~$\scale\in\intFromTo{16}{256}$} have been selected based on a large experiment in~\cite{WCLW2019SADSOBIFATMPFBBDOA}, where they are described in detail.
They are also used in~\cite{WWLC2020FFAMOAIUBTOTOFV}, where the \opofeaP\ performed much better on them than the \opoeaP.

Expanding on this prior finding, we can see from Figure~\ref{fig:wmodel_results} that \opoeaP\ and \ggaP\ have a 100\% success rate only on four \wmodel\ instances.
The \sagaP\ can always solve the first five and has more than 80\% success rate on the sixth instance.
It can solve most of the \wmodel\ instances more often than the other pure algorithms (with the exception of instances~9 and~11).
In the top part of the figure, we plot the empirically estimated expected runtime (\ert)~\cite{HAFR2012RPBBOBES}.
We find that \sagaP\ tends to have a lower \ert\ on the harder instances while being slower on the easier ones compared to the other pure algorithms.

All algorithms with FFA can solve all \wmodel\ instances in all runs within the budget.
With the exception of instance~3, their mean runtimes are below or approximately equal to the third power of the instance scale~\scale.
All of them also have similar \ert\ values which, again with the exception of instance~3, are always better than those of all pure algorithms.%
\subsection{\maxsat\ Problems}%
\begin{table}[tb]%
\caption{The number of failed runs (out of 11'000) over the instance scales~$\scale$.}%
\label{tbl:maxsat:success}%
\centering%
\includegraphics[width=0.99\linewidth]{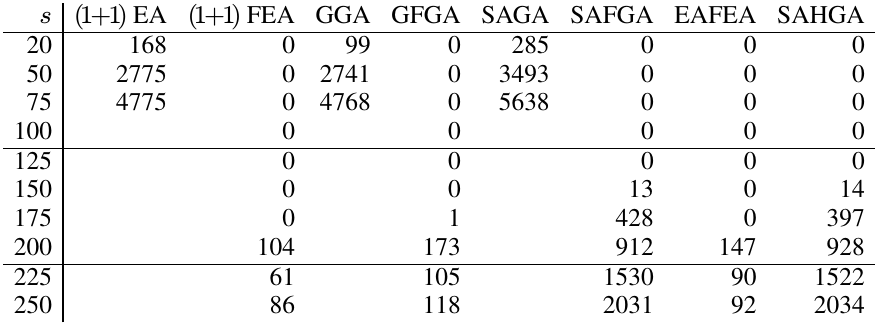}%
\end{table}%
\begin{table}[tb]%
\caption{The logarithm base~10 of the \ert\ over 11'000~runs on each instance scale~$\scale$.}%
\label{tbl:maxsat:ert}%
\centering%
\includegraphics[width=0.99\linewidth]{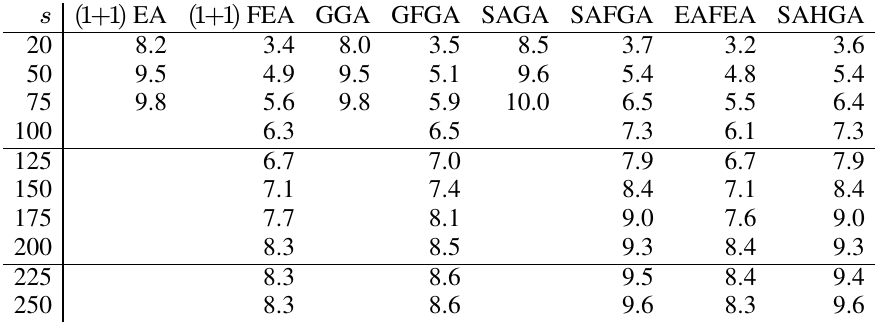}%
\end{table}%
A \maxsat\ instance is a formula~\mbox{$\maxSatFormula:\booleans^{\scale}\rightarrow\booleans$} over~\scale\ Boolean variables~\cite{HS2005SLSFAA}.
The variables appear as literals either directly or negated in~\maxSatClauses\ ``{\texttt{or}}'' clauses, which are all combined into one ``{\texttt{and}}''.
The objective function \objFunb{\solspel}, subject to minimization, computes the number of clauses which are \texttt{false} under the variable setting~\solspel.
If~\mbox{$\objFunb{\solspel}=0$}, all clauses are \texttt{true}, which solves the problem.
The \maxsat\ problem exhibits low epistasis but deceptiveness~\cite{RW1998GABITMD}.
In the so-called phase transition region with~\mbox{$\maxSatClauses/\maxSatVariables\approx4.26$}, the average instance hardness for stochastic local search algorithms is maximal~\cite{HS2000SAORFROS,DNS2017TCAOEAORSkCF,DNS2015IRBFTOPOEOR3CFBOFDC}.

We apply our algorithms as incomplete solvers~\cite{GKSS2008SS} on the ten sets of \emph{satisfiable} uniform random \mbox{3-SAT} instances from SATLib~\cite{HS2000SAORFROS}, which stem from this region.
Here, the number of variables~\scale\ is \mbox{from~$\left\{20\right\}\cup\left\{25i:i\in \intFromTo{2}{10}\right\}$}, where 1000~instances are given for \mbox{$\scale\in\left\{20,50,100\right\}$} and 100 otherwise.
For each scale level, we conduct 11'000~runs with our algorithms exactly uniformly distributed over all available instances.
With the pure algorithms, we can only do this for \mbox{$\scale\in\left\{20,50,75\right\}$} due to the high runtime requirement resulting from many runs failing to solve the problem within \mbox{$10^{10}$~FEs}.

From Table~\ref{tbl:maxsat:success}, we can see that the highest number of failed runs at scale~$\scale=250$ of \emph{any} algorithm using FFA is lower than the lowest number of failed runs of \emph{any} pure algorithm at~$\scale=50$.
From Table~\ref{tbl:maxsat:ert}, we find that no FFA-based algorithm has a higher \ert\ at scale~$\scale=250$ than its pure variant on~$\scale=50$.
On the scales~$\scale\leq 75$, the FFA-based algorithms have a mean runtime which is between 3 and 4 orders of magnitude smaller that the \ert\ of the pure algorithms.
Overall, the \opofeaP\ performs best, closely followed by the \eafeaP.
These results provide striking evidence regarding the suitability of FFA to solve \npHard\ and practically relevant problems.%
\subsection{Summary}%
\begin{table}[tb]%
\caption{The values~$t$ corresponding to upper limits~$\scale^t$ of the runtime in FEs never exceeded by any run. $\emptyset$ $\equiv$ some runs failed or were not conducted due to infeasible runtime.}%
\label{tbl:summary}%
\centering%
\includegraphics[width=0.99\linewidth]{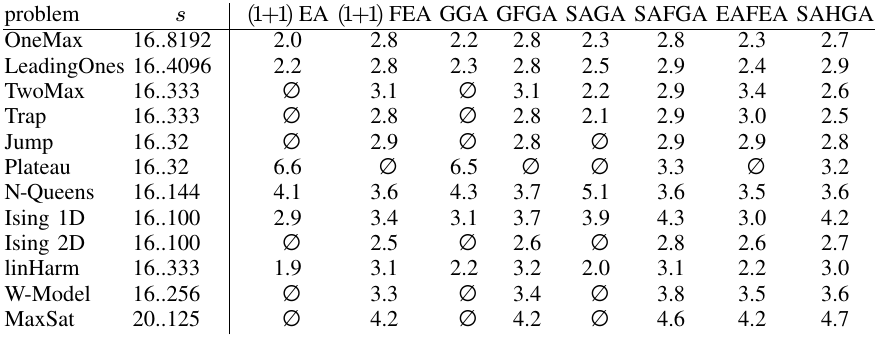}%
\end{table}%
In Table~\ref{tbl:summary}, we summarize our experimental results.
We try to give an impression on the worst-case time that the algorithms needed to solve the problems in relation to the problem scale~\scale, regardless of any other problem parameter.
We computed~$t=\max\{\log_{\scale}{\runtime_{\scale}}\}$, looking for values~$t$ for which~$\scale^t$ marks an upper limit for the worst runtimes~$\runtime_{\scale}$ in FEs any run has consumed for solving any instance of the problem at a scale~\scale.

On some problems, some of the runs were not successful or instances were skipped because it was clear that they would not finish within the budget. 
Both situation are signified with~$\emptyset$.
All columns with numerical values indicate that all runs on all explored scales in the range were successful.

From the table, we see again that \safgaP\ and \safgapP\ are the only algorithms that always succeed on all 11~problems on the listed scales.
For them, the \isingod\ is the hardest among the theory-inspired benchmarks.
On the seven problem types that the \sagaP\ can solve efficiently, its FFA-based variants only win on \nqueens.

The performance of the \sagaP\ suffers when its population size~$\lambda$ becomes unnecessarily large, e.g., on the \leadingones, \nqueens, \isingod, and \maxsat\ problems.
Limiting~$\lambda$ to not exceed a constant like~5 or~10 or to a scale-dependent value such as~$2\ln{\scale}$ could be a remedy for this issue~\cite{BD2017OSMSDFTOPLCLEAOO}.
It would likely improve the performance of the three variants of the \sagaP\ on problems with weak fitness distance correlation.
In turn, it would probably remove the ability of the pure \sagaP\ to efficiently solve the \twomax\ and \trap\ problems and the ability of the \safgaP\ and \safgapP\ to efficiently solve \plateau\ problems.
Alternative approaches would be applying self-adaptation with rollbacks~\cite{BBS2021TOFRWRFSAOTPSITOPLCLGA}, resetting~$\lambda$ to~1 when it reaches~$\scale$ and still cannot improve upon the best-so-far solution~\cite{HFS2020OTCOTPCMITOPLCLGA}, or choosing~$\lambda$ using a power-law distribution in each iteration~\cite{ABD2020FMICBA}.

The \opofeaP, \eafeaP, and \gfgaP\ can reliably solve all theory-inspired problems except {\plateau}s.
The pure algorithms have problems with {\jump}s, {\plateau}s, \isingtd, and some \wmodel\ instances.
Interestingly, \isingtd\ appears to be easy for FFA-based methods.
The FFA-based algorithms can reliably solve the \npHard\ \maxsat\ problem within the budget up to $\scale=125$ and have their largest \mbox{$t$-values} on this problem.
The good performance of the \opoeaP\ relative to the \ggaP\ and \sagaP\ in Table~\ref{tbl:summary} can be explained -- at least on some problems -- with the relatively low scales~{\scale} we investigated, with the fact that we use worst-case runtimes, and with our somewhat crude approach, which boils down performance to only a monomial, disregarding possible lower-order polynomials as well as constant or logarithmic factors.%
\section{Conclusions}%
\label{sec:conclusions}%
Frequency Fitness Assignment (FFA) is an approach to prevent premature convergence of optimization processes.
It has three very interesting and unique already known properties:
(1)~it creates optimization processes that are \emph{not} biased towards better solutions~\cite{WLCW2021SJSSPWUABFGS};
(2)~it renders them invariant under all bijective transformations of the objective function value~\cite{WWLC2020FFAMOAIUBTOTOFV};
(3)~it can reduce the empirical mean runtime of the \opoeaP\ on several benchmark problems from exponential to polynomial in our experiments~\cite{WWLC2020FFAMOAIUBTOTOFV}.

In this paper, we significantly extended the knowledge along the lines of the third property above.
We introduced FFA into two state-of-the-art EAs for discrete optimization, namely the \ggaP\ and the \sagaP.
On problems that these algorithms can already solve efficiently, the addition of FFA leads to a slow-down which often seems to be linear in the number of distinct objective values.

On several of the investigated benchmark problems where \ggaP\ and \opoeaP\ need exponential expected runtime (\twomax, \trap, \jump), the \gfgaP, \opofeaP, and the hybrid \eafeaP\ seem to exhibit polynomial mean runtime.
However, they retain the exponential mean runtime of their pure variants on the \plateau\ problems.
On \jump\ and \plateau, where the \sagaP\ also exhibits exponential expected runtime, the \safgaP\ with FFA and the hybrid \safgapP\ seem to require polynomial mean runtime only.
They are also much faster on the \nqueens\ problem where the \sagaP\ is slow.
It is notable and surprising that FFA does not degenerate the self-adjusting method of \sagaP.
It even interacts with it positively, allowing it to solve \emph{all} benchmark problems except \maxsat\ efficiently.

We furthermore investigated two ways to hybridize FFA with direct optimization (\safgapP, \eafeaP) and found that both performed very well.
For example, on \nqueens\ and \isingtd, the hybrid \eafeaP\ was the fastest algorithm in our study.

With this article, we have confirmed that FFA is not an oddity that only, somehow, makes a basic and practically unimportant \opoeaP\ more efficient on hard problems.
We found that it can also do so with two of the best EAs for discrete black-box optimization available.

The \safgaP\ and \safgapP\ were not necessarily the best algorithms in the lot, in particular on the \maxsat\ and \isingod\ problems.
Nevertheless, the existence of algorithms that may be able to solve all investigated benchmark problems from theory efficiently is very motivating.
We will therefore study more variants~\cite{BD2017OSMSDFTOPLCLEAOO,BBS2021TOFRWRFSAOTPSITOPLCLGA,ABD2020FMICBA} of these algorithms.

We believe that there is much potential in better hybrids of FFA and pure optimization.
The good performance of the \eafeaP\ on \isingtd\ is another indicator for this.
We will also follow this research direction in the future.

Our first results for the \opofeaP\ on the Job Shop Scheduling Problem~\cite{WLCW2021SJSSPWUABFGS} are very encouraging, too.
We will therefore extend the scope of our experiments to more \npHard\ optimization tasks and other representations.%
%
%\newpage%
\section*{Acknowledgments}%
We acknowledge support by %
the National Natural Science Foundation of China under Grant 61673359, % Weise Thomas
the Hefei Specially Recruited Foreign Expert support, % Weise Thomas
the Key Research Plan of Anhui~202104d07020006 and 2022k07020011, % Wu Zhize
the Key Common Technology and Major Scientific Achievement Engineering Project of Hefei~2021GJ030, % Wu Zhize
the Youth Project of the Provincial Natural Science Foundation of Anhui~1908085QF285, % Wu Zhize
and %
the University Natural Sciences Research Projects of Anhui Province%
~KJ2019A0835 and % Li Xinlu
KJ2020A0661.% Wu Zhize

We are deeply thankful for the feedback and help of the anonymous reviewers that greatly helped us improving the article.
In particular, we want to thank the reviewer who pointed out that our FFA makes algorithms invariant under \emph{injective} transformations of the objective function value.
In~\cite{WWLC2020FFAMOAIUBTOTOFV} and the original version of this article, we stated that the invariance requires the transformation function~$g$ to be \emph{bijective}, but it is indeed already sufficient if it is injective.%
\ifCLASSOPTIONcaptionsoff%
\newpage%
\fi%
\bibliographystyle{IEEEtran}%

\begin{thebibliography}{10}%
\providecommand{\url}[1]{#1}
\csname url@samestyle\endcsname
\providecommand{\newblock}{\relax}
\providecommand{\bibinfo}[2]{#2}
\providecommand{\BIBentrySTDinterwordspacing}{\spaceskip=0pt\relax}
\providecommand{\BIBentryALTinterwordstretchfactor}{4}
\providecommand{\BIBentryALTinterwordspacing}{\spaceskip=\fontdimen2\font plus
\BIBentryALTinterwordstretchfactor\fontdimen3\font minus
  \fontdimen4\font\relax}
\providecommand{\BIBforeignlanguage}[2]{{%
\expandafter\ifx\csname l@#1\endcsname\relax
\typeout{** WARNING: IEEEtran.bst: No hyphenation pattern has been}%
\typeout{** loaded for the language `#1'. Using the pattern for}%
\typeout{** the default language instead.}%
\else
\language=\csname l@#1\endcsname
\fi
#2}}
\providecommand{\BIBdecl}{\relax}
\BIBdecl

\bibitem{WWTWDY2014FFA}
T.~Weise, M.~Wan, K.~Tang, P.~Wang, A.~Devert, and X.~Yao, ``Frequency fitness
  assignment,'' \emph{{IEEE} Trans.\ Evol.\ Comput.}, vol.~18, no.~2, pp.
  226--243, 2014.

\bibitem{WLCW2021SJSSPWUABFGS}
T.~Weise, X.~Li, Y.~Chen, and Z.~Wu, ``Solving job shop scheduling problems
  without using a bias for good solutions,'' in \emph{Genetic and Evol.\
  Comput.\ Conf.\ Companion ({GECCO'21})}.\hskip 1em plus 0.5em minus
  0.4em\relax ACM, 2021.

\bibitem{WWLC2020FFAMOAIUBTOTOFV}
T.~Weise, Z.~Wu, X.~Li, and Y.~Chen, ``Frequency fitness assignment: Making
  optimization algorithms invariant under bijective transformations of the
  objective function value,'' \emph{{IEEE} Trans.\ Evol.\ Comput.}, vol.~25,
  pp. 307--319, 2021.

\bibitem{WWTY2014EEIAWGP}
T.~Weise, M.~Wan, K.~Tang, and X.~Yao, ``Evolving exact integer algorithms with
  genetic programming,'' in \emph{{IEEE} Congress on Evol.\ Comput.\
  ({CEC'14})}.\hskip 1em plus 0.5em minus 0.4em\relax Los Alamitos, CA, USA:
  {IEEE}, 2014, pp. 1816--1823.

\bibitem{WCLW2019SADSOBIFATMPFBBDOA}
T.~Weise, Y.~Chen, X.~Li, and Z.~Wu, ``Selecting a diverse set of benchmark
  instances from a tunable model problem for black-box discrete optimization
  algorithms,'' \emph{Appl.\ Soft Comput.}, vol.~92, p. 106269, 2019.

\bibitem{S2012CSUBBA}
D.~Sudholt, ``Crossover speeds up building-block assembly,'' in \emph{Genetic
  and Evol.\ Comput.\ Conf.\ {(GECCO'12)}}.\hskip 1em plus 0.5em minus
  0.4em\relax {ACM}, 2012, pp. 689--702.

\bibitem{S2017HCSUBBAIGA}
------, ``How crossover speeds up building block assembly in genetic
  algorithms,'' \emph{Evol.\ Comput.}, vol.~25, no.~2, pp. 237--274, 2017.

\bibitem{DDE2015FBBCTDNGA}
B.~Doerr, C.~Doerr, and F.~Ebel, ``From black-box complexity to designing new
  genetic algorithms,'' \emph{Theor.\ Comput.\ Sci.}, vol. 567, pp. 87--104,
  2015.

\bibitem{DD2018OSASAPCFT1LLGA}
B.~Doerr and C.~Doerr, ``Optimal static and self-adjusting parameter choices
  for the $(1+(\lambda,\lambda))$ genetic algorithm,'' \emph{Algorithmica},
  vol.~80, no.~5, pp. 1658--1709, 2018.

\bibitem{CPD2017TAMPARAOEA}
\BIBentryALTinterwordspacing
E.~{Carvalho Pinto} and C.~Doerr, ``Towards a more practice-aware runtime
  analysis of evolutionary algorithms,'' 2017, \mbox{arXiv:1812.00493v1}
  \mbox{[cs.NE]} \mbox{3~Dec~2018}. [Online]. Available:
  \url{http://arxiv.org/pdf/1812.00493.pdf}
\BIBentrySTDinterwordspacing

\bibitem{DYHWSB2020BDOHWI}
C.~Doerr, F.~Ye, N.~Horesh, H.~Wang, O.~M. Shir, and T.~B{\"a}ck,
  ``Benchmarking discrete optimization heuristics with \mbox{IOHprofiler},''
  \emph{Appl.\ Soft Comput.}, vol.~88, p. 106027, 2020.

\bibitem{GR1987GAS}
D.~E. Goldberg and J.~T. Richardson, ``Genetic algorithms with sharing for
  multimodal function optimization,'' in \emph{Intl.\ Conf.\ on Genetic
  Algorithms and their Applications ({ICGA'87})}.\hskip 1em plus 0.5em minus
  0.4em\relax Mahwah, NJ, USA: Lawrence Erlbaum Associates, Inc., 1987, pp.
  41--49.

\bibitem{WWZXLW2021HODFVHRSI}
Z.~Wu, X.~Wang, L.~Zou, L.~Xu, X.~Li, and T.~Weise, ``Hierarchical object
  detection for very high-resolution satellite images,'' \emph{Appl. Soft
  Comput.}, vol. 113, p. 107885, 2021.

\bibitem{OAAH2017IGOAAUPVIP}
Y.~Ollivier, L.~Arnold, A.~Auger, and N.~Hansen, ``Information-geometric
  optimization algorithms: {A} unifying picture via invariance principles,''
  \emph{J.\ Mach.\ Learn.\ Res.}, vol.~18, pp. 18:1--18:65, 2017.

\bibitem{CLOW2021OSSEAASPWIRBAORPIB}
D.~Corus, A.~Lissovoi, P.~S. Oliveto, and C.~Witt, ``On steady-state
  evolutionary algorithms and selective pressure: Why inverse rank-based
  allocation of reproductive trials is best,'' \emph{{ACM} Trans. Evol. Learn.
  Optim.}, vol.~1, no.~1, pp. 2:1--2:38, 2021.

\bibitem{GLY2019QDTS}
D.~Gravina, A.~Liapis, and G.~N. Yannakakis, ``Quality diversity through
  surprise,'' \emph{{IEEE} Trans.\ Evol.\ Comput.}, vol.~23, no.~4, pp.
  603--616, 2019.

\bibitem{CD2018QADOAUMF}
A.~Cully and Y.~Demiris, ``Quality and diversity optimization: {A} unifying
  modular framework,'' \emph{{IEEE} Trans.\ Evol.\ Comput.}, vol.~22, no.~2,
  pp. 245--259, 2018.

\bibitem{PSS2016QDANFFEC}
J.~K. Pugh, L.~B. Soros, and K.~O. Stanley, ``Quality diversity: {A} new
  frontier for evolutionary computation,'' \emph{Frontiers Robotics {AI}},
  vol.~3, p.~40, 2016.

\bibitem{LS2011AOETTSFNA}
J.~Lehman and K.~O. Stanley, ``Abandoning objectives: Evolution through the
  search for novelty alone,'' \emph{Evol.\ Comput.}, vol.~19, no.~2, pp.
  189--223, 2011.

\bibitem{GLY2016SSBOAN}
D.~Gravina, A.~Liapis, and G.~N. Yannakakis, ``Surprise search: Beyond
  objectives and novelty,'' in \emph{Genetic and Evol.\ Comput.\ Conf.\
  ({GECCO'16})}.\hskip 1em plus 0.5em minus 0.4em\relax {ACM}, 2016, pp.
  677--684.

\bibitem{CPD2018ASPFTUOCIBBO}
E.~{Carvalho Pinto} and C.~Doerr, ``A simple proof for the usefulness of
  crossover in black-box optimization,'' in \emph{Parallel Problem Solving from
  Nature {(PPSN~XV)}, Part {II}}.\hskip 1em plus 0.5em minus 0.4em\relax
  Springer, 2018, pp. 29--41.

\bibitem{DD2015OPCTSAAT15TRIDS}
B.~Doerr and C.~Doerr, ``Optimal parameter choices through self-adjustment:
  Applying the 1/5-th rule in discrete settings,'' in \emph{Genetic and Evol.\
  Comput.\ Conf.\ ({GECCO'15})}.\hskip 1em plus 0.5em minus 0.4em\relax {ACM},
  2015, pp. 1335--1342.

\bibitem{M1992HGARWMAH}
H.~M{\"{u}}hlenbein, ``How genetic algorithms really work: Mutation and
  hillclimbing,'' in \emph{Parallel Problem Solving from Nature~2
  (\mbox{PPSN-II})}.\hskip 1em plus 0.5em minus 0.4em\relax Elsevier, 1992, pp.
  15--26.

\bibitem{DJW2006UALBFRSHIBBO}
S.~Droste, T.~Jansen, and I.~Wegener, ``Upper and lower bounds for randomized
  search heuristics in black-box optimization,'' \emph{Theory Comput.\ Syst.},
  vol.~39, no.~4, pp. 525--544, 2006.

\bibitem{W1989TGAASPWRBAORTIB}
L.~D. Whitley, ``The {GENITOR} algorithm and selection pressure: Why rank-based
  allocation of reproductive trials is best,'' in \emph{Intl.\ Conf.\ on
  Genetic Algorithms ({ICGA'89})}.\hskip 1em plus 0.5em minus 0.4em\relax San
  Francisco, CA, USA: Morgan Kaufmann Publishers Inc., 1989, pp. 116--121.

\bibitem{R1997CPOEA}
G.~Rudolph, \emph{Convergence Properties of Evolutionary Algorithms}.\hskip 1em
  plus 0.5em minus 0.4em\relax Hamburg, Germany: Verlag Dr.\ Kova{\v{c}}, 1997.

\bibitem{AADLMW2013TQCOFAHP}
P.~Afshani, M.~Agrawal, B.~Doerr, K.~G. Larsen, K.~Mehlhorn, and C.~Winzen,
  ``The query complexity of finding a hidden permutation,'' in
  \emph{Space-Efficient Data Structures, Streams, and Algorithms}.\hskip 1em
  plus 0.5em minus 0.4em\relax Springer, 2013, ch.~1, pp. 1--11.

\bibitem{AADDLM2019TQCOAPBVOM}
P.~Afshani, M.~Agrawal, B.~Doerr, C.~Doerr, K.~G. Larsen, and K.~Mehlhorn,
  ``The query complexity of a permutation-based variant of mastermind,''
  \emph{Discret.\ Appl.\ Math.}, vol. 260, pp. 28--50, 2019.

\bibitem{DJW2002OTAOTOPOEA}
S.~Droste, T.~Jansen, and I.~Wegener, ``On the analysis of the \mbox{$(1+1)$}
  evolutionary algorithm,'' \emph{Theor.\ Comput.\ Sci.}, vol. 276, no. 1-2,
  pp. 51--81, 2002.

\bibitem{ADK2019ATRAFTOPLCLGOL}
D.~Antipov, B.~Doerr, and V.~Karavaev, ``A tight runtime analysis for the
  $(1+(\lambda,\lambda))$ {GA} on {LeadingOnes},'' in \emph{15th {ACM/SIGEVO}
  Conference on Foundations of Genetic Algorithms ({FOGA'19})}.\hskip 1em plus
  0.5em minus 0.4em\relax {ACM}, 2019, pp. 169--182.

\bibitem{KAD2019TAESOTOPLCLEOTLP}
V.~Karavaev, D.~Antipov, and B.~Doerr, ``Theoretical and empirical study of the
  $(1 + (\lambda,\lambda))$~{EA} on the {LeadingOnes} problem,'' in
  \emph{Genetic and Evol.\ Comput.\ Conf.\ Companion {(GECCO'19)}}.\hskip 1em
  plus 0.5em minus 0.4em\relax {ACM}, 2019, pp. 2036--2039.

\bibitem{FQW2018ELDBOAWHTMO}
T.~Friedrich, F.~Quinzan, and M.~Wagner, ``Escaping large deceptive basins of
  attraction with heavy-tailed mutation operators,'' in \emph{Genetic and
  Evol.\ Comput.\ Conf.\ ({GECCO'18})}.\hskip 1em plus 0.5em minus 0.4em\relax
  {ACM}, 2018, pp. 293--300.

\bibitem{VHGN2002FTTIMEAHSP}
C.~{Van Hoyweghen}, D.~E. Goldberg, and B.~Naudts, ``From {TwoMax} to the
  {Ising} model: Easy and hard symmetrical problems,'' in \emph{Genetic and
  Evol.\ Comput.\ Conf.\ ({GECCO'02})}.\hskip 1em plus 0.5em minus 0.4em\relax
  Morgan Kaufmann, 2002, pp. 626--633.

\bibitem{FOSW2009AODPMFGE}
T.~Friedrich, P.~S. Oliveto, D.~Sudholt, and C.~Witt, ``Analysis of
  diversity-preserving mechanisms for global exploration,'' \emph{Evol.\
  Comput.}, vol.~17, no.~4, pp. 455--476, 2009.

\bibitem{NB2003AAOTBOSEAOTF}
S.~Nijssen and T.~B{\"{a}}ck, ``An analysis of the behavior of simplified
  evolutionary algorithms on trap functions,'' \emph{{IEEE} Trans.\ Evol.\
  Comput.}, vol.~7, no.~1, pp. 11--22, 2003.

\bibitem{QGWF2021EAASFBOHTM}
F.~Quinzan, A.~G{\"{o}}bel, M.~Wagner, and T.~Friedrich, ``Evolutionary
  algorithms and submodular functions: benefits of heavy-tailed mutations,''
  \emph{Natural Computing}, vol.~20, no.~3, pp. 561--575, 2021.

\bibitem{DDK2015UBBCOJF}
B.~Doerr, C.~Doerr, and T.~K{\"{o}}tzing, ``Unbiased black-box complexities of
  jump functions,'' \emph{Evol.\ Comput.}, vol.~23, no.~4, pp. 641--670, 2015.

\bibitem{AD2021PRAFPF}
D.~Antipov and B.~Doerr, ``Precise runtime analysis for plateau functions,''
  \emph{{ACM} Trans.\ Evol.\ Learn.\ Optim.}, vol.~1, no.~4, pp. 13:1--13:28,
  2021.

\bibitem{FW2005TODIMMVR}
S.~Fischer and I.~Wegener, ``The one-dimensional {Ising} model: Mutation versus
  recombination,'' \emph{Theor.\ Comput.\ Sci.}, vol. 344, no. 2-3, pp.
  208--225, 2005.

\bibitem{F2004APUBFAMBAOTTDIM}
S.~Fischer, ``A polynomial upper bound for a mutation-based algorithm on the
  two-dimensional {Ising} model,'' in \emph{Genetic and Evol.\ Comput.\ Conf.\
  {(GECCO'04)}}.\hskip 1em plus 0.5em minus 0.4em\relax Springer, 2004, pp.
  1100--1112.

\bibitem{WW2018DFOCOPATTWMBPFST}
T.~Weise and Z.~Wu, ``Difficult features of combinatorial optimization problems
  and the tunable {\wmodel} benchmark problem for simulating them,'' in
  \emph{Companion Mat.\ Genetic and Evol.\ Comput.\ Conf.\ ({GECCO'18})}.\hskip
  1em plus 0.5em minus 0.4em\relax {ACM}, 2018, pp. 1769--1776.

\bibitem{WNSRG2008ATMFMOERANFL}
T.~Weise, S.~Niemczyk, H.~Skubch, R.~Reichle, and K.~Geihs, ``A tunable model
  for multi-objective, epistatic, rugged, and neutral fitness landscapes,'' in
  \emph{Annual Conf.\ on Genetic and Evol.\ Comput.\ ({GECCO'08})}.\hskip 1em
  plus 0.5em minus 0.4em\relax New York, {USA}: {ACM} Press, 2008, pp.
  795--802.

\bibitem{HAFR2012RPBBOBES}
\BIBentryALTinterwordspacing
N.~Hansen, A.~Auger, S.~Finck, and R.~Ros, ``Real-parameter black-box
  optimization benchmarking: Experimental setup,'' Universit{\'{e}} Paris Sud,
  INRIA Futurs, {\'{E}}quipe TAO, Orsay, France, Tech. Rep., 2012. [Online].
  Available:
  \url{http://coco.lri.fr/BBOB-downloads/download11.05/bbobdocexperiment.pdf}
\BIBentrySTDinterwordspacing

\bibitem{HS2005SLSFAA}
H.~H. Hoos and T.~St{\"{u}}tzle, \emph{Stochastic Local Search: Foundations and
  Applications}.\hskip 1em plus 0.5em minus 0.4em\relax San Francisco, CA, USA:
  Morgan Kaufmann Publishers Inc., 2005.

\bibitem{RW1998GABITMD}
S.~B. Rana and L.~D. Whitley, ``Genetic algorithm behavior in the {MAXSAT}
  domain,'' in \emph{Parallel Problem Solving from Nature ({PPSN~V})}.\hskip
  1em plus 0.5em minus 0.4em\relax Springer, 1998, pp. 785--794.

\bibitem{HS2000SAORFROS}
H.~H. Hoos and T.~St{\"{u}}tzle, ``{SATLIB}: An online resource for research on
  {SAT},'' in \emph{{SAT2000} -- Highlights of Satisfiability Research in the
  Year 2000}.\hskip 1em plus 0.5em minus 0.4em\relax Amsterdam, The
  Netherlands: {IOS} Press, 2000, pp. 283--292.

\bibitem{DNS2017TCAOEAORSkCF}
B.~Doerr, F.~Neumann, and A.~M. Sutton, ``Time complexity analysis of
  evolutionary algorithms on random satisfiable \mbox{k-CNF} formulas,''
  \emph{Algorithmica}, vol.~78, no.~2, pp. 561--586, 2017.

\bibitem{DNS2015IRBFTOPOEOR3CFBOFDC}
------, ``Improved runtime bounds for the \mbox{(1+1) EA} on random
  \mbox{3-CNF} formulas based on fitness-distance correlation,'' in
  \emph{Genetic and Evol.\ Comput.\ Conf.\ ({GECCO'15})}.\hskip 1em plus 0.5em
  minus 0.4em\relax {ACM}, 2015, pp. 1415--1422.

\bibitem{GKSS2008SS}
C.~P. Gomes, H.~A. Kautz, A.~Sabharwal, and B.~Selman, ``Satisfiability
  solvers,'' in \emph{Handbook of Knowledge Representation}.\hskip 1em plus
  0.5em minus 0.4em\relax Elsevier, 2008, pp. 89--134.

\bibitem{BD2017OSMSDFTOPLCLEAOO}
M.~V. Buzdalov and B.~Doerr, ``Runtime analysis of the $(1+(\lambda,\lambda))$
  genetic algorithm on random satisfiable {3-CNF} formulas,'' in \emph{Genetic
  and Evol.\ Comput.\ Conf.\ {(GECCO'17)}}.\hskip 1em plus 0.5em minus
  0.4em\relax {ACM}, 2017, pp. 1343--1350.

\bibitem{BBS2021TOFRWRFSAOTPSITOPLCLGA}
A.~O. Bassin, M.~V. Buzdalov, and A.~A. Shalyto, ``The ``one-fifth rule'' with
  rollbacks for self-adjustment of the population size in the
  $(1+(\lambda,\lambda))$ genetic algorithm,'' \emph{Autom.\ Control.\ Comput.\
  Sci.}, vol.~55, no.~7, pp. 885--902, 2021.

\bibitem{HFS2020OTCOTPCMITOPLCLGA}
M.~A. {Hevia Fajardo} and D.~Sudholt, ``On the choice of the parameter control
  mechanism in the $(1+(\lambda\,\lambda))$ genetic algorithm,'' in
  \emph{Genetic and Evol.\ Comput.\ Conf.\ ({GECCO'20})}.\hskip 1em plus 0.5em
  minus 0.4em\relax {ACM}, 2020, pp. 832--840.

\bibitem{ABD2020FMICBA}
D.~Antipov, M.~V. Buzdalov, and B.~Doerr, ``Fast mutation in crossover-based
  algorithms,'' in \emph{Genetic and Evol.\ Comput.\ Conf.\
  ({GECCO'20})}.\hskip 1em plus 0.5em minus 0.4em\relax {ACM}, 2020, pp.
  1268--1276.

\end{thebibliography}
\newpage%
\begin{IEEEbiography}[{\includegraphics[width=1in,height=1.25in,keepaspectratio]{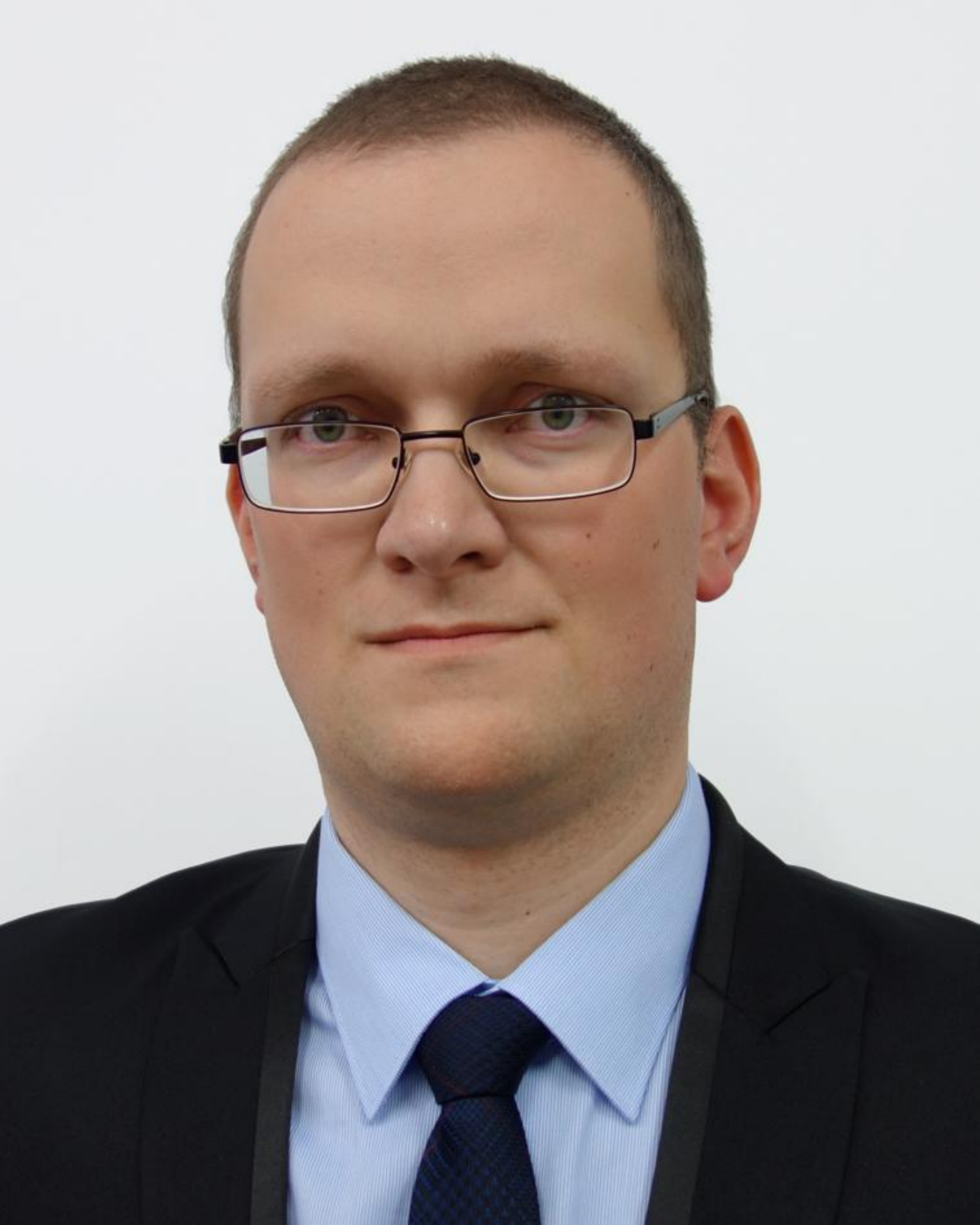}}]{Thomas Weise} obtained his Ph.D.\ in Computer Science from the University of Kassel (Kassel, Germany) in 2009.
He then joined the University of Science and Technology of China (Hefei, Anhui, China), first as PostDoc, later as Associate Professor.
In 2016, he moved to Hefei University (in Hefei) as Full Professor to found the Institute of Applied Optimization (IAO) at the School of Artificial Intelligence and Big Data.
His research interests include metaheuristic optimization and algorithm benchmarking and he is author or co-author of 115 peer-reviewed academic publications, including
over 40~articles and 65~conference papers.
He was awarded the title \emph{Hefei Specially Recruited Foreign Expert} of the city Hefei in Anhui, China in~2019 and the Hefei City Friendship Award in~2020.%
\end{IEEEbiography}%
\vspace{-1cm}%
\begin{IEEEbiography}[{\includegraphics[width=1in,height=1.25in,keepaspectratio]{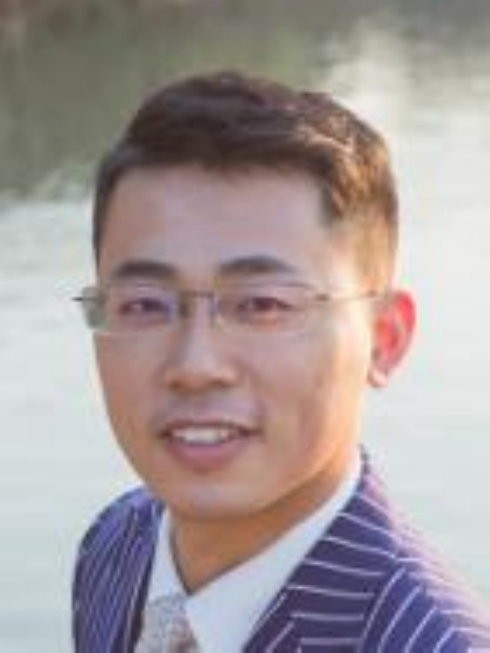}}]{Zhize Wu} obtained his Bachelor from the Anhui Normal University (Wuhu, Anhui, China) in 2012 and his Ph.D.\ in Computer Science from the University of Science and Technology of China (Hefei, Anhui, China) in 2017.
He then joined the Institute of Applied Optimization (IAO) of the School of Artificial Intelligence and Big Data at Hefei University as Lecturer.
His research interest include  image processing, neural networks, deep learning, machine learning in general, and algorithm benchmarking.%
\end{IEEEbiography}%
\vspace{-1cm}%
\begin{IEEEbiography}[{\includegraphics[width=1in,height=1.25in,clip,keepaspectratio]{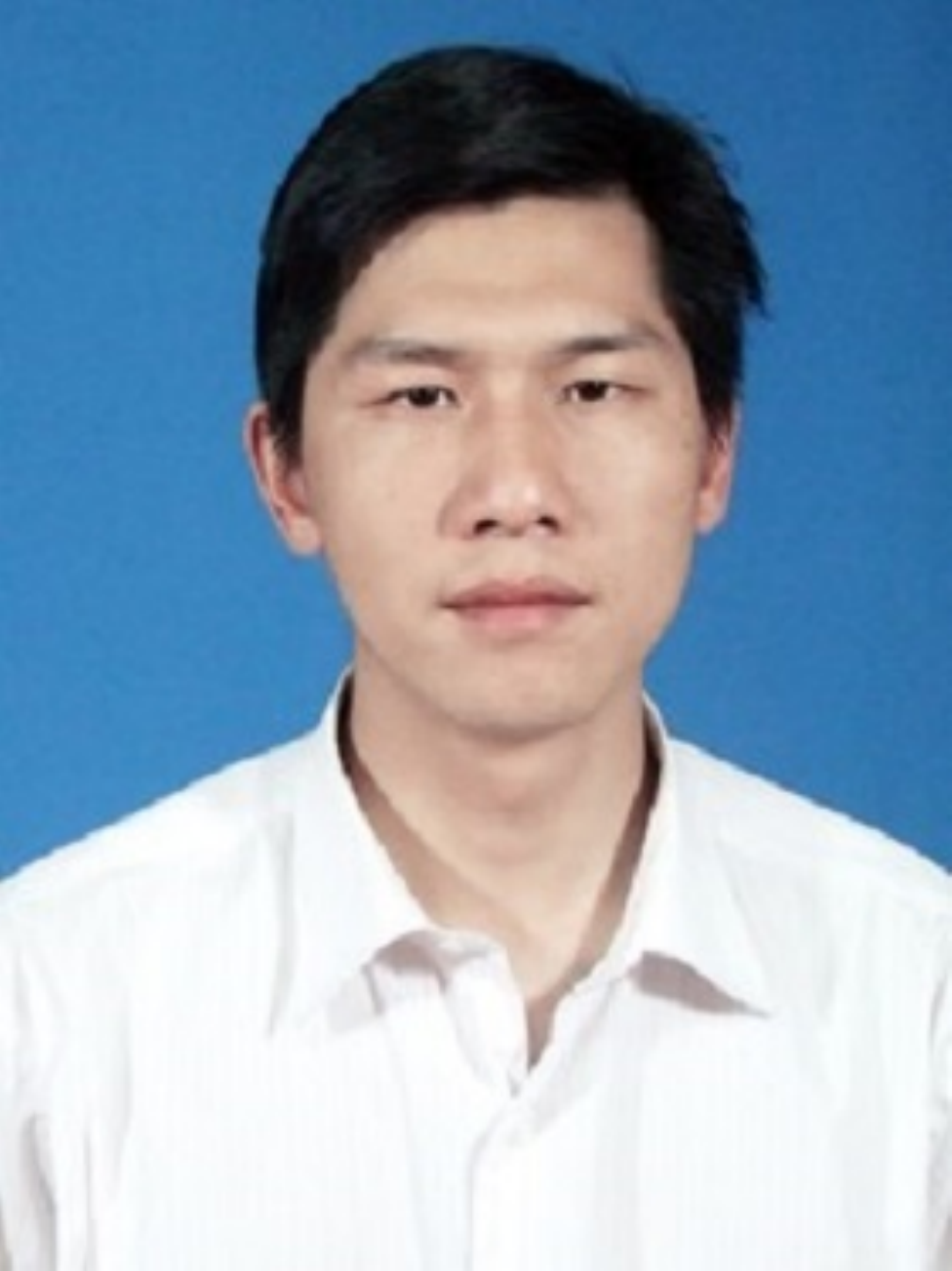}}]{Xinlu Li} received his Bachelor from the Fuyang Normal University (Fuyang, Anhui, China) in 2002, his M.Sc.~degree in Computer Science from Anhui University (Hefei, Anhui, China) in 2009, and his Ph.D.~degree in Computer Science from the TU~Dublin (Dublin, Ireland) in 2019.
His career at Hefei University started as Lecturer in 2009, Senior Lecturer in 2012, in 2018 he joined the Institute of Applied Optimization (IAO), and in 2019 he became Associate Researcher.
From 2014 to 2018, he was Teaching Assistant in the School of Computing of TU~Dublin.
His research interest include Swarm Intelligence and optimization, algorithms which he applies to energy efficient routing protocol design for large-scale Wireless Sensor Networks.%
\end{IEEEbiography}%
\vspace{-1cm}%
\begin{IEEEbiography}[{\includegraphics[width=1in,height=1.25in,keepaspectratio]{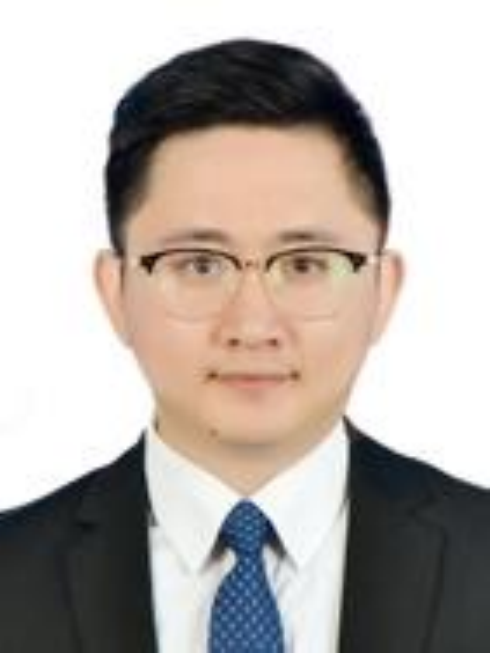}}]{Yan Chen} received his Bachelor from the Anhui University of Science and Technology (Huai'nan, Anhui, China) in 2011, his M.Sc.~degree from the China University of Mining and Technology (Xuzhou, Jiangsu, China) in 2014, and his doctorate from the TU Dortmund (Dortmund, Germany) in 2019.
There, he studied spatial information management and modeling as a member of the Spatial Information Management and Modelling Department of the School of Spatial Planning.
He joined the Institute of Applied Optimization (IAO) of the School of Artificial Intelligence and Big Data at Hefei University as Lecturer in 2019.
His research interests include hydrological modeling, Geographic Information System, Remote Sensing, and Sponge City/Smart City applications, as well as deep learning and optimization algorithms.%
\end{IEEEbiography}%
\vspace{-1cm}%
\begin{IEEEbiography}[{\includegraphics[width=1in,height=1.25in,keepaspectratio]{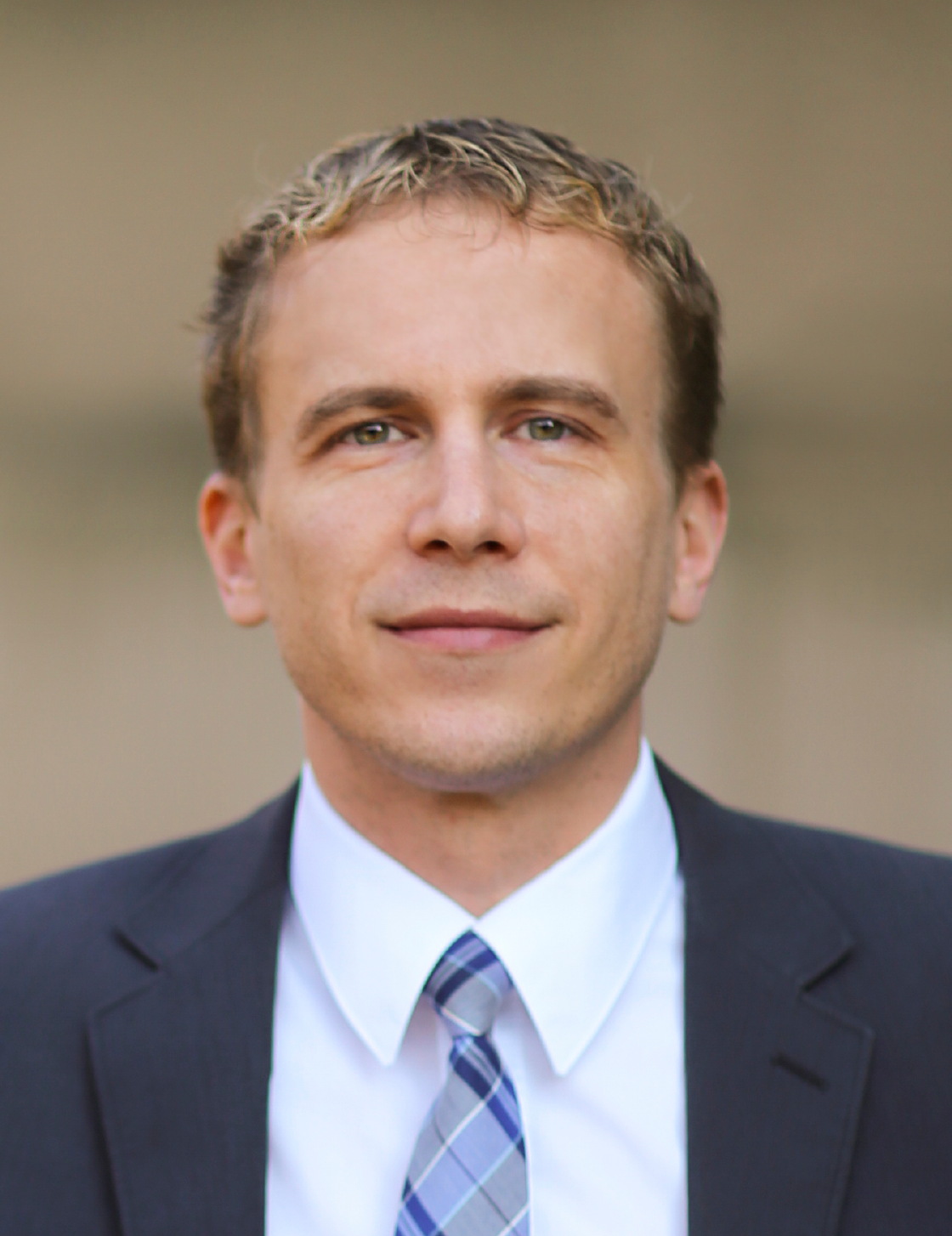}}]{J{\"o}rg L{\"a}ssig} is Full Professor for enterprise application development at the University of Applied Sciences Zittau/G{\"o}rlitz, Germany and also Research Manager at the Fraunhofer Society.
After completing his studies in computer science and computational physics, he received his PhD in 2009 from the Chemnitz University of Technology on efficient algorithms and models for cooperation generation and control.
This was followed by postdoctoral stays at the International Computer Science Institute in Berkeley, California and at the Universit{\'a} della Svizzera italiana in Lugano, Switzerland.
His research includes metaheuristics for optimization, quantum algorithmics and cybersecurity for critical infrastructures.%
\end{IEEEbiography}%
\end{document}